\documentclass[10pt,twocolumn,letterpaper]{article}

\usepackage{cvpr}
\usepackage{times}
\usepackage{epsfig}
\usepackage{graphicx}
\usepackage{amsmath}
\usepackage{amssymb}
\usepackage{placeins}
\usepackage{subcaption}
\usepackage{tabularx}
\usepackage{multirow}
\usepackage{hhline}
\usepackage{arydshln}
\usepackage{graphbox}
\usepackage[export]{adjustbox}
\usepackage{enumitem}
\usepackage{array}
\newcolumntype{M}[1]{>{\centering\arraybackslash}m{#1}}
\usepackage{booktabs, makecell}

\newlength{\oldtabcolsep}

\newcommand{\fig}[1]{Figure~\ref{fig:#1}}
\newcommand{\sect}[1]{Section~\ref{sect:#1}}
\newcommand{\tab}[1]{Table~\ref{tab:#1}}

\usepackage[pagebackref=true,breaklinks=true,letterpaper=true,colorlinks,bookmarks=false]{hyperref}

\cvprfinalcopy 

\def\cvprPaperID{8178} 

\ifcvprfinal\fi
\begin{document}

\title{Neural Head Reenactment with Latent Pose Descriptors}

\author{
Egor Burkov$^{1,2}$
\quad
Igor Pasechnik$^{1}$
\quad
Artur Grigorev$^{1}$
\quad
Victor Lempitsky$^{1,2}$
\vspace{1.5mm}\\
$^1$Samsung AI Center, Moscow \quad $^2$Skolkovo Institute of Science and Technology
}

\maketitle
\begin{abstract}
We propose a neural head reenactment system, which is driven by a latent pose representation and is capable of predicting the foreground segmentation alongside the RGB image. The latent pose representation is learned as a part of the entire reenactment system, and the learning process is based solely on image reconstruction losses. We show that despite its simplicity, with a large and diverse enough training dataset, such learning successfully decomposes pose from identity. The resulting system can then reproduce mimics of the driving person and, furthermore, can perform cross-person reenactment. Additionally, we show that the learned descriptors are useful for other pose-related tasks, such as keypoint prediction and pose-based retrieval.
\end{abstract}

\section{Introduction}

\newlength{\widei}
\setlength{\widei}{0.12\textwidth}
\addtolength{\tabcolsep}{-5pt}
\begin{figure*}
    \centering
    \begin{tabular}{cccccccc}
    \includegraphics[clip,trim=0 0 256px 0,width=\widei]{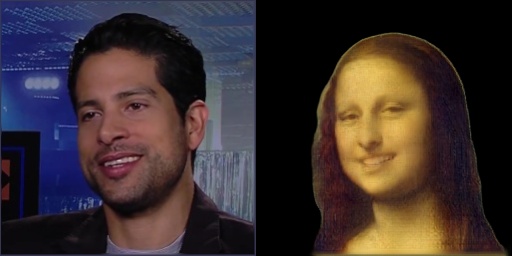} &
    \includegraphics[clip,trim=0 0 256px 0,width=\widei]{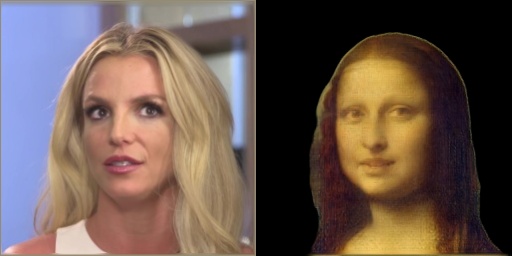} &
    \includegraphics[clip,trim=0 0 256px 0,width=\widei]{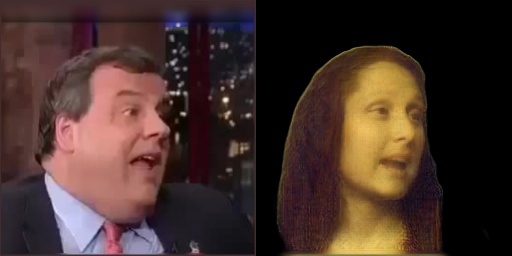} &
    \includegraphics[clip,trim=0 0 256px 0,width=\widei]{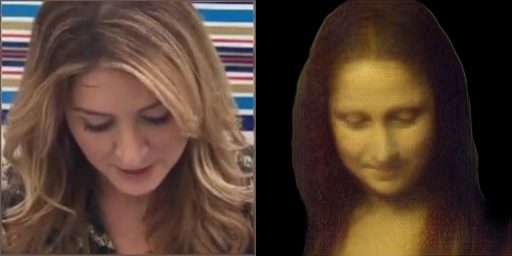} &
    \includegraphics[clip,trim=0 0 256px 0,width=\widei]{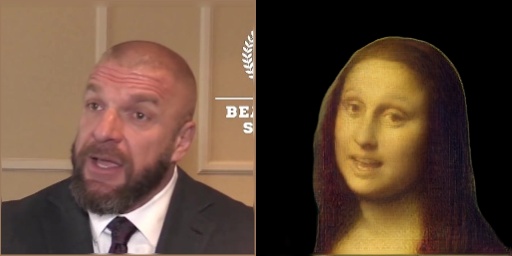} &
    \includegraphics[clip,trim=0 0 256px 0,width=\widei]{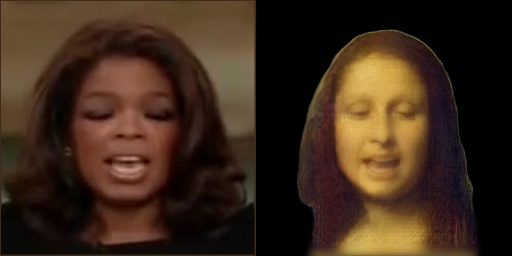} &
    \includegraphics[clip,trim=0 0 192px 0,width=\widei]{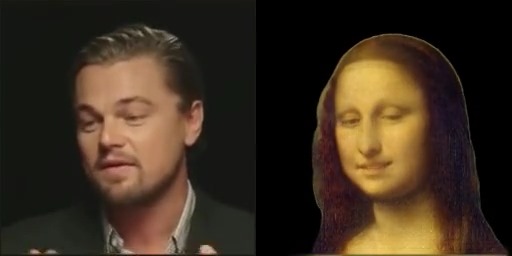} &
    \includegraphics[clip,trim=0 0 192px 0,width=\widei]{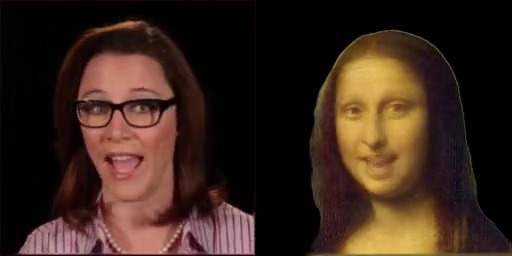} \\
    \includegraphics[clip,trim=256px 0 0 0,width=\widei]{figures/teaser/id00061_cAT9aR8oFx0_00002.jpg} &
    \includegraphics[clip,trim=256px 0 0 0,width=\widei]{figures/teaser/id01228_7qHTvs0VO68_00000.jpg} &
    \includegraphics[clip,trim=256px 0 0 0,width=\widei]{figures/teaser/id01593_p1wo17ZRcFw_00301.jpg} &
    \includegraphics[clip,trim=256px 0 0 0,width=\widei]{figures/teaser/id07802_BfQUBDw7TiM_00028.jpg} &
    \includegraphics[clip,trim=256px 0 0 0,width=\widei]{figures/teaser/id08701_UeUyLqpLz70_00013.jpg} &
    \includegraphics[clip,trim=256px 0 0 0,width=\widei]{figures/teaser/id06692_t8M_6FT07Co_00201.jpg} &
    \includegraphics[clip,trim=192px 0 0 0,width=\widei]{figures/teaser/7.jpg} &
    \includegraphics[clip,trim=192px 0 0 0,width=\widei]{figures/teaser/6.jpg}
    \end{tabular}
     \caption{\textbf{Being a Mona Lisa.} Our system can generate realistic reenactments of arbitrary talking heads (such as Mona Lisa) using arbitrary people as pose drivers (top row). Despite learning in an unsupervised setting, the method can successfully decompose pose and identity, so that the identity of the reenacted person is preserved. }
    \label{fig:teaser}
\end{figure*}

\addtolength{\tabcolsep}{5pt}

Head video reenactment has seen dramatic progress in quality and robustness over the recent years. Current state-of-the-art systems \cite{Suwajanakorn17,Kim18,Wang18c,Pumarola18,Wiles18a,Zakharov19,Fu19,Wang19,Tripathy19} demonstrate compelling nearly photorealistic ``talking head'' reenactments. The most recent ones are able to accomplish this even when a single image of the target person is available~\cite{Pumarola18,Zakharov19,Fu19,Wang19,Tripathy19}, by using deep neural generative networks. 
In this work, we improve on the pre-existing neural one-shot head reenactment systems in two important ways. First, rather straightforwardly, we augment the previous state-of-the-art reenactment system~\cite{Zakharov19} with the ability to predict foreground segmentation. Such prediction is needed for various scenarios, such as telepresence, where the transfer of the original background to the new environment can be undesirable. 

More importantly, we suggest a new pose\footnote{Here and below, by `head pose' we mean the combination of head orientation, position, as well as facial expression.} representation for neural head reenactment. The representation of the pose plays the key role in the quality of reenactment. Most systems, including~\cite{Suwajanakorn17,Kim18,Wang18c,Zakharov19,Fu19,Wang19}, are based on keypoint (landmark) representation. The main advantage of such representation is that robust and efficient ``off-the-shelf'' landmark detectors are now available~\cite{Kazemi14,Bulat17}. 

Face landmarks, however, suffer from several shortcomings. First, learning a landmark detector requires excessive annotation effort, and the sets of annotated landmarks often miss some important aspects of the pose. E.g.\ many landmark annotations do not include eye pupils, and as a consequence, the reenactment will not have a full control of the gaze. Second, many of the landmarks do not have an anatomical basis, and their annotation is ambiguous and prone to errors, especially, when they are occluded. In practice, such ambiguity of annotation often translates into temporal instability of keypoint detection that in turn translates into the reenactment results. Finally, as a representation, landmarks are \emph{person-specific}, as they contain considerable amount of information about pose-independent head geometry. This may be highly undesirable for head reenactment, e.g.\ if one wants to drive an iconic photograph or painting with the target person having a different head geometry.

The emergence of large unlabeled datasets of human videos such as \cite{Nagrani17,Chung18,Chung17} allows to learn \textit{latent} pose-expression descriptors in an unsupervised way. This approach has been first explored in \cite{Wiles18a,Wiles18b}, where the latent pose descriptors were learned such that the dense flow between different frames can be inferred from the learned descriptors. In this work, we propose an alternative to the warping-based approach~\cite{Wiles18a,Wiles18b}. Our approach learns low-dimensional \textit{person-agnostic} pose descriptors alongside with medium-dimensional \textit{person-specific} pose-independent descriptors, by imposing a set of reconstruction losses on video frames over a large collection of videos. Importantly, when evaluating the reconstruction losses, we segment out the background, so that the background clutter and its change across frames does not affect the learned descriptors.

We show that a simple learning framework based on sampling multiple random frames from the same video paired with the large size of the video dataset allows to learn extractors for both descriptors that work very well for reenactment tasks, including cross-person reenactment. In particular, we show that our reenactment based on the new latent pose representation preserves the identity of the target person much better than when FAb-Net~\cite{Wiles18b} and X2Face~\cite{Wiles18a} pose descriptors are used.
Additionally, we analyze the quality of learned latent pose descriptors for such tasks as landmark prediction and pose-based retrieval.

\section{Related work}
\label{sect:related}

Face/head reenactment is an active area of research. Here we distinguish between works where changes and augmentations are localized within faces (face reenactment), e.g.~\cite{Suwajanakorn17,Pumarola18}, and more ambitious approaches that model extended regions including significant portion of clothing, neck, upper garment (head reenactment), e.g.~\cite{Kim18,Wang18c,Zakharov19}.

Pose representation is an important aspect of reenactment systems. As mentioned above, most works drive reenactment using landmarks \cite{Suwajanakorn17,Kim18,Wang18c,Zakharov19,Fu19,Wang19}.  Another approach is to use facial action units (AU)~\cite{Ekman77}, as is done in face reenactment~\cite{Pumarola18} and head reenactment~\cite{Tripathy19}. Detecting action units still requires manual annotation and supervised learning. The X2Face system~\cite{Wiles18a} uses latent vectors that are learned to be predictable of warping fields.

A more classic approach is to model face/head pose in the 3D morphable model (3DMM) framework~\cite{Blanz99} or using a similar approach in 2D (e.g.\ an active appearance model)~\cite{Cootes01}. Still, learning 3DMM and fitting a learned 3DMM almost invariably involves detecting landmarks, thus inheriting many of the landmark deficiencies. Alternatively, a dataset of 3D scans is required to build a model for pose/identity disentanglement in 3DMM framework.

Several recent works have investigated how landmarks can be learned in an unsupervised way~\cite{Zhang18,Jakab18}. While generally very promising, unsupervised keypoints still contain person-specific information just like supervised keypoints, and therefore are not generally suitable for cross-person reenactment. Same applies to dense, high-dimensional descriptors such as DensePose body descriptor~\cite{Guler18}, and dense face-only descriptors~\cite{Guler17,Thewlis19}. Finally, Codec avatars~\cite{Lombardi18} learn person-specific latent pose descriptors and extractors based on the reconstruction losses. However, the transfer of such descriptors from person to person was not considered. The recent and parallel work~\cite{Siarohin19} has demonstrated that \textit{relative} motion of unsupervised keypoints can be used to transfer animations at least in the absence of strong head rotation. Full-fledged comparison of our approach to~\cite{Siarohin19} is left for future work.

Beyond head/face reenactment, there is a very large body of work on learning disentangled representations. Some representative works that learn latent pose or shape descriptors for arbitrary classes of objects using datasets of videos include \cite{Denton17,Xiao19}. Some approaches (e.g.~\cite{Liu19}) aim to learn content-style disentanglement (which may roughly correspond to shape-texture disentanglement) using adversarial~\cite{Goodfellow14} and cycle-consistency~\cite{Zhu17,Huang18} losses. Alternatively, disentanglement can be obtained by the direct fitting of factorized distributions to data (e.g.~\cite{Kim18b}).

\section{Method}
\label{sect:method}

\begin{figure*}[!hbt]
    \centering
    \includegraphics[width=0.8\textwidth]{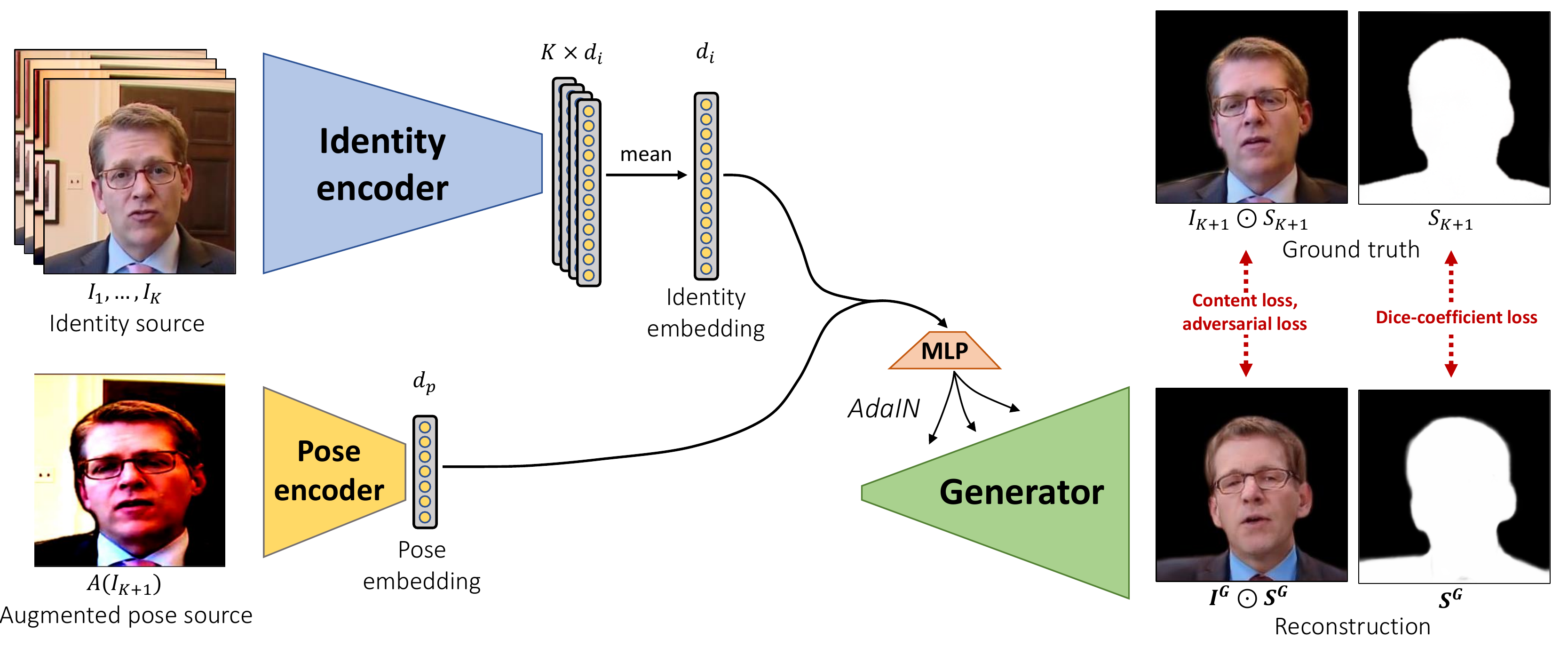}
    \caption{At each step of meta-learning, our system samples a set of frames from a video of a person. The frames are processed by two encoders. The bigger \textit{identity encoder} is applied to several frames of the video, while the smaller \textit{pose encoder} is applied to a hold-out frame. The obtained embeddings are passed to the generator network, whose goal is to reconstruct the last (hold-out) frame. Since the capacity of the pose encoder is limited, and since its input does not exactly match other frames w.r.t. identity (thanks to data augmentation), the system learns to extract all pose-independent information through the identity encoder, and uses the smaller encoder to capture only pose-related information, thus achieving pose-identity disentanglement.}
    \label{fig:pipeline}
\end{figure*}

Our system modifies and expands the reenactment model of Zakharov~et~al.~\cite{Zakharov19}. First, the ability to predict the segmentation is added. Second, the system learns to perform reenactment based on latent pose vectors rather than keypoints. Below, we provide the details of our system (shown in \fig{pipeline}).

\subsection{Training pipeline}

As in \cite{Zakharov19}, we learn on the VoxCeleb2 dataset~\cite{Chung18} of video sequences. Each sequence contains a talking person, and is obtained from a raw sequence by running a face detector, cropping the resulting face and resizing it to a fixed size ($256 \times 256$ in our case). Also, as in the case of \cite{Zakharov19}, there is a ``meta-learning'' stage when a big model responsible for reproducing all people in the dataset is trained through a sequence of training episodes, and a fine-tuning stage, when that ``meta-model'' is fine-tuned to a tuple of images (or a single image) of a particular person. Our changes affect both stages in an analogous way, and we focus on the meta-learning step in the below discussion.

During each episode of meta-learning we consider a single video sequence. We then fetch $K+1$ random frames $I_1, \dots, I_{K+1}$ from this sequence, as well as $S_{K+1}$ ~-- a foreground segmentation map for $I_{K+1}$, which we precompute using an off-the-shelf semantic segmentation network. The first $K$ images $I_1, \dots, I_K$ are then fed into a relatively high-capacity convolutional net $F$, which we call \textit{identity encoder}. It is analogous to the embedder network in \cite{Zakharov19} with the exception that it does not accept the keypoints as an input. 

For each image $I_k$, the identity encoder outputs $d_i$-dimensional vector $x_k=F(I_k)$, which we call the \textit{identity embedding} of $I_k$. Identity embeddings are expected to contain the pose-independent information about the person (including lighting, clothing, etc.) Given $K$ frames, we obtain a single identity vector $\bar x$ by taking the mean of $x_1,..x_K$. 

The remaining image $I_{K+1}$ (the \textit{pose source}) first undergoes a random \textit{pose augmentation} transformation $A$, which is described below. Then, $A(I_{K+1})$ is passed through a network of much lower capacity, which we call the \textit{pose encoder} and denote as $G$. The pose encoder outputs a $d_p$-dimensional \textit{pose embedding} $y_{K+1}=G(A(I_{K+1}))$, which we want to be a \textit{person-agnostic} pose descriptor.

The transformation $A$ mentioned above is important for pose-identity disentanglement. It keeps person's pose intact but may alter its identity. Namely, it randomly scales the image independently over the horizontal and the vertical axes, and randomly applies content-preserving operations such as blur, sharpening, contrast change, or JPEG compression. We call $A$ \textit{pose augmentation} since it is applied on the pose source, and it can be regarded as a form of data augmentation.

The pose and the identity embeddings are passed to the \textit{generator network} that tries to reconstruct the image $I_{K+1}$ as accurately as possible. Whereas \cite{Zakharov19} used rasterized keypoints (stickman images) to pass the pose into their generator networks, we rely entirely on the AdaIN~\cite{Huang17} mechanism to pass both the pose and the identity embeddings to the generator. More specifically, our upsampling generator starts with a constant learnable tensor of size $512 \times 4 \times 4$ and outputs the two tensors: $I^G(\bar x, y_{K+1})$ of size $3 \times 256 \times 256$ and $S^G(\bar x, y_{K+1})$ of size $1 \times 256 \times 256$, which it tries to match to the foreground part of the image $I_{K+1}$ and its segmentation mask $S_{K+1}$ respectively. This is achieved by simply predicting a $4 \times 256 \times 256$ tensor in the final layer. The AdaIN blocks are inserted after each convolution. The AdaIN coefficients are produced by taking the concatenated pose and identity embeddings and passing this $(d_i+d_p)$-dimensional vector through an MLP with learnable parameters in the spirit of StyleGAN~\cite{Karras19}.

We expect $I^G(\bar x,y_{K+1}) \odot S^G(\bar x,y_{K+1})$ and $S^G(\bar x,y_{K+1})$ produced by the generator to be as close as possible to $I_{K+1} \odot S_{K+1}$ and $S_{K+1}$, respectively. We achieve this with the help of several loss functions. Segmentation maps are matched with the help of the dice coefficient loss \cite{Milletari16}. Head images with background blacked out, on the other hand, are matched using the same combination of losses as in \cite{Zakharov19}. Namely, there are content losses based on matching of ConvNet activations for a VGG-19 model trained for ImageNet classification and a VGGFace model trained for face recognition.
Also, $I^G \odot S^G$ and $I_{K+1} \odot S_{K+1}$ are passed through a projection discriminator (the difference from \cite{Zakharov19} here is that we again do not provide rasterized keypoints to it) to compute the adversarial loss that pushes images to be realistic, the discriminator feature matching loss, and an embedding match term. 

\paragraph{Reenactment and fine-tuning.} Once the model has been meta-learned, it can be used to fit new identities unseen during meta-learning. Thus, given one or more images of a new person, their identity vector $\bar x$ can be extracted by passing those images through the identity encoder and averaging the results element-wise. Then, by plugging in a pose vector $y$ extracted from an image of the same or of a different person, we can reenact the person by computing the image $I^G(\bar x,y)$ and its foreground mask $S^G(\bar x,y)$. 

To further reduce the identity gap, we follow \cite{Zakharov19} and fine-tune the model (namely, the weights of the MLP, the generator, and the discriminator) with the same set of losses as in \cite{Zakharov19} plus the dice coefficient loss treating the provided set of images of a new person and their segmentation as the ground truth. The estimated identity embedding $\bar x$ is kept fixed during the fine-tuning (including it into the optimization did not result in any difference in our experiments, since the number of parameters in the embedding $\bar x$ is much smaller than in the MLP and the generator network). The pose embedding network $G$ is also kept fixed during the fine-tuning.

\subsection{Disentanglement of pose and identity}
\label{sect:disentanglement}

Our key finding is that when applied to a person $X$ the reenactment model trained as discussed above can successfully reproduce the mimics of a person in image $I$ when the pose vector $y=G(I)$ is extracted from an image of the same person $X$. More surprisingly, the model can also reproduce the mimics when the pose vector is extracted from an image of a different person $Y$. In this case, the \textit{bleeding of identity} from this different person is kept to a minimum, i.e.\ the resulting image still looks like an image of a person $X$.

Initially, we expected that such disentanglement of pose and identity should not happen, and that some form of adversarial training~\cite{Goodfellow14} or cycle-consistency~\cite{Zhu17,Huang18} would be necessary to ensure the disentanglement. It turns out that with (i) low enough capacity of the pose extractor network $G$, (ii) pose augmentations applied, and (iii) background segmented out, disentanglement happens automatically, and our experiments with extra loss terms such as in e.g.~\cite{Denton17} did not produce any further improvement. Apparently, with the three techniques above, the model prefers to extract all person-specific details from the identity source frame using the higher-capacity identity extractor network.

In \sect{results} below, we evaluate this disentanglement effect that came as a ``pleasant surprise'' and show that it is indeed stronger than in the case of other related approaches (i.e.~supports cross-person reenactment better with less identity bleeding).

In \textbf{supplementary material}, we additionally conduct ablation studies to investigate how pose encoder capacity, pose augmentations, segmentation, and latent pose vector dimensionality $d_p$ affect the ability of our reenactment system to preserve pose and identity.

\subsection{Implementation details}

Our training dataset is a collection of YouTube videos from VoxCeleb2~\cite{Chung18}.
There are on the order of 100,000 videos of about 6,000 people. We sampled 1 of every 25 frames from each video, leaving around seven million of total training images.
In each image, we re-cropped the annotated face by first capturing its bounding box with the S$^3$FD detector \cite{Zhang17}, then making that box square by enlarging the smaller side, growing the box's sides by $80\%$ keeping the center, and finally resizing the cropped image to $256 \times 256$.
Human segmentation is obtained by the Graphonomy model \cite{Gong19}.
As in \cite{Zakharov19}, we set $K=8$ thus using the identity vector extracted from eight random frames of a video in order to reconstruct the ninth one based on its pose descriptor.

In our best model, the pose encoder has the MobileNetV2 architecture~\cite{Sandler18}  and the identity encoder is a ResNeXt-50 (32 $\times$ 4d)~\cite{Xie17}.
Both have not been tweaked, and so they include batch normalization \cite{Ioffe15}. The pose and identity embedding sizes, $d_p$ and $d_i$, are 256 and 512 respectively. No normalization or regularization is applied to the embeddings. The module that transforms them into AdaIN parameters is a ReLU perceptron with spectral normalization and one hidden layer of 768 neurons.

Our generator is based on that of \cite{Zakharov19}, but without donwsampling blocks, since all inputs are delegated to AdaINs, which are located after each convolution. More precisely, a $512 \times 4 \times 4$ learnable constant tensor is transformed by 2 constant resolution residual blocks, followed by 6 upsampling residual blocks. We start halving the number of channels from the fourth upsampling block so that the tensor of final resolution ($256 \times 256$) has 64 channels. That tensor is passed through an AdaIN layer, a ReLU, a $1 \times 1$ convolution and a $\tanh$, becoming a 4-channel image. Unlike \cite{Zakharov19}, we do not use self-attention. Spectral normalization \cite{Miyato18b} is employed everywhere in the generator, the discriminator and the MLP.

Instead of alternating generator and discriminator updates, a single weight update is carried out for all networks after gradient accumulations from all loss terms.

We trained the model for 1,200,000 iterations with a minibatch of 8 samples spread over two NVIDIA P40 GPUs, which in total takes about two weeks.

\section{Results}
\label{sect:results}

Our quantitative evaluation assesses both the relative performance of the pose descriptors using auxiliary tasks, and the quality of cross-person reenactment. Qualitatively, we show examples of reenactment in same-person and cross-person scenarios as well as interpolation results in the learned pose space. The ablation study in the supplementary material shows the effect of different components of our method.

\subsection{Compared methods}
\label{sect:compared-methods}

Below, we compare our results with the results of the following methods and systems. We consider the following pose descriptors based on various degrees of supervision:

\begin{itemize}

\item \textbf{Ours.} 256-dimensional latent pose descriptors learned within our system.

\item \textbf{X2Face.} 128-dimensional \textit{driving vectors} learned within the X2Face reenactment system~\cite{Wiles18a}.

\item \textbf{FAb-Net.} We also evaluate the 256-dimensional FAb-Net descriptors~\cite{Wiles18b} as a pose representation. These are related to ours in that, although not being person-agnostic, they are also learned in an unsupervised way from the VoxCeleb2 video collection.

\item \textbf{3DMM.} We consider a state-of-the-art 3DMM system~\cite{Chang18}. This system extracts decomposed rigid pose, face expression, and a shape descriptor using a deep network. The pose descriptor is obtained by concatenating the rigid pose rotation (represented as a quaternion), and the face expression parameters (29 coefficients).

\end{itemize}

Our descriptor learns from VoxCeleb2 dataset. X2Face descriptor is trained on a smaller VoxCeleb1 dataset~\cite{Nagrani17}, and FAb-Net is learned from both. The 3DMM descriptors are most heavily supervised, as the 3DMM is learned from 3D scans and requires a landmark detector (which is in turn learned in a supervised setting).

In addition, we consider the following head reenactment systems based on these pose descriptors:
\begin{itemize}

\item \textbf{\textsf{Ours}}. Our full system as described in \sect{method}.

\item \textbf{\textsf{X2Face}}. The X2Face system~\cite{Wiles18a} based on native descriptors and warping-based reenactment.

\item \textbf{\textsf{X2Face+}}. In this variant, we use frozen pre-trained X2Face's \textit{driving network} (up to the \textit{driving vector}) instead of our pose encoder, and keep the rest of the architecture unchanged from ours. We train the identity encoder, the generator conditioned on X2Face latent pose vector and our identity embedding, and the projection discriminator.

\item \textbf{\textsf{FAb-Net+}}. Same as \textsf{X2Face+} but with frozen FAb-Net in place of our pose encoder.

\item \textbf{\textsf{3DMM+}}. Same as \textsf{X2Face+} but with frozen ExpNet~\cite{Chang18} in place of our pose encoder, and with pose augmentations disabled. The pose descriptor is constructed from ExpNet's outputs as described above. We additionally normalize these 35-dimensional descriptors by per-element mean and standard deviation computed over the VoxCeleb2 training set.

\item \textbf{\textsf{FSTH}}. The original few-shot talking head system of \cite{Zakharov19} driven by rasterized keypoints.

\item \textbf{\textsf{FSTH+}}. We retrain the system of \cite{Zakharov19} by making several changes that makes it more comparable with our system and other baselines. The raw keypoint coordinates are put into the generator using AdaIN mechanism (just like in our system). The generator predicts segmentation alongside the image. We also use the same crops, which are different from \cite{Zakharov19}.
\end{itemize}

\subsection{Descriptor evaluation}

\begin{table}[]
\centering
\begin{tabular}{l|c|c|c|c}

\multicolumn{5}{c}{\textbf{Accuracy for Top-N queries (\%)}} \\ \hline
\textbf{Descriptor} & \multicolumn{1}{c}{\textbf{N=10}} & \multicolumn{1}{c}{\textbf{N=20}} & \multicolumn{1}{c}{\textbf{N=50}} & \multicolumn{1}{c}{\textbf{N=100}} \\ \hline
FAb-Net & \multicolumn{1}{c}{45.7} & \multicolumn{1}{c}{40.8} & \multicolumn{1}{c}{36.6} & \multicolumn{1}{c}{35.7} \\ \hline
3DMM & \multicolumn{1}{c}{47.3} & \multicolumn{1}{c}{45.6} & \multicolumn{1}{c}{41.9} & \multicolumn{1}{c}{41.1} \\ \hline
X2Face & \multicolumn{1}{c}{61.0} & \multicolumn{1}{c}{55.7} & \multicolumn{1}{c}{51.8} & \multicolumn{1}{c}{49.4} \\ \hline
\textbf{Ours} & \multicolumn{1}{c}{\textbf{75.7}} & \multicolumn{1}{c}{\textbf{63.8}} & \multicolumn{1}{c}{\textbf{57.8}} & \multicolumn{1}{c}{\textbf{54.1}}
\end{tabular}
\caption{The accuracy of pose (expression)-based retrieval results using different pose descriptors on the Multi-PIE dataset. See text for more details.}
\label{tab:retrieval}
\end{table}

To understand how good are the learned pose descriptors at matching different people in the same pose, we use the Multi-PIE dataset~\cite{gross2008}, which is not used for training of either of descriptors, but has six emotion class annotations for people in various poses. We restrict the dataset to near-frontal and half-profile camera orientations (namely \texttt{08\_0, 13\_0, 14\_0, 05\_1, 05\_0, 04\_1, 19\_0}), leaving 177,280 images. In each camera orientation group, we randomly choose a query image from it and fetch the closest $N$ images from the same group using cosine similarity of descriptors. We consider a match to be correct if a person with the same emotion label is returned. We repeat this procedure 100 times for each group. In \tab{retrieval} we show the overall ratio of correct matches within top-10, top-20, top-50, and top-100 lists. For the 3DMM descriptor, we only consider the 29 face expression coefficients and ignore the rigid pose information as irrelevant for emotions.

In this comparison, it can be observed that the latent space of our pose embeddings is better grouped with respect to emotion classes that those of other facial expression descriptors, as our result is much better for top-10 and top-20 metrics, while being similar to X2Face and better than the rest for top-50 and top-100. We believe that FAb-Net's and X2Face's vectors contain identity information (we prove this again below), so they are more likely to be close to vectors representing same or similar person. As for 3DMM, it requires different latent expression vectors to turn different shapes (persons) into the same facial expression by construction; therefore, expression coefficients may easily coincide for different people showing different facial expressions.

\paragraph{Keypoint prediction.} Keypoint regression is not within our target applications, since keypoints contain person-specific information. However, this is a popular task on which unsupervised pose descriptors have compared in the past, so we have also run our method on a standard benchmark on the MAFL \cite{Liu2015} test set. To predict keypoints, we use a ReLU MLP with one hidden layer of size 768, and in our case we use both pose and identity embeddings as an input. Using the standard normalized inter-ocular distance, we obtain the distance error of 2.63. This is smaller than the error of 3.44 obtained by FAb-Net, though behind the state-of-the-art of \cite{Jakab18} (2.54) for this task.

\addtolength{\tabcolsep}{-4pt}
\setlength{\widei}{0.53\textwidth}
\begin{figure}
    \centering
    \includegraphics[clip,trim=35 35 0 65,width=\widei]{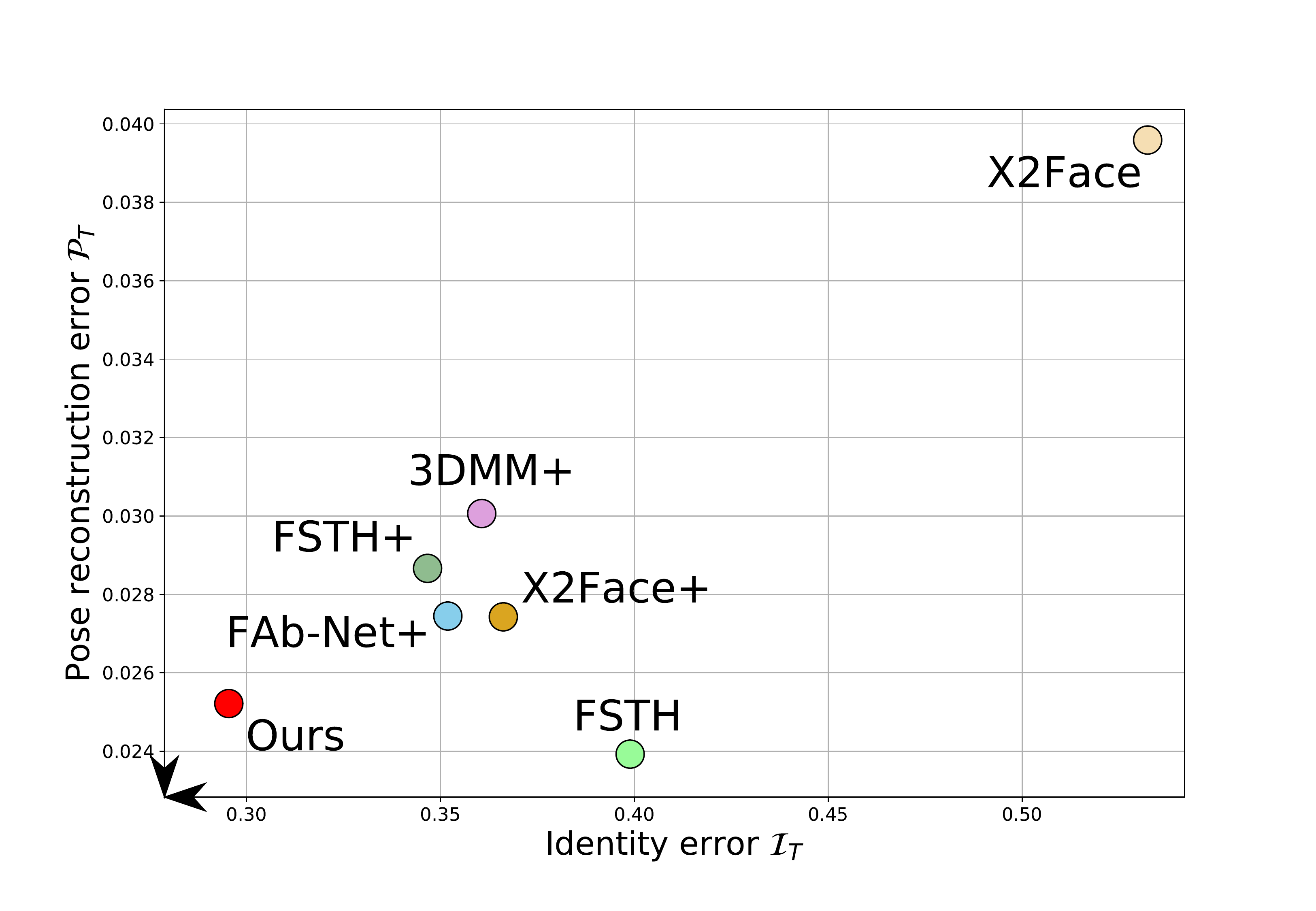}
    \caption{Evaluation of reenactment systems in terms of their ability to represent the driver pose and to preserve reference identity (arrows point towards improvement). See text for details.}
    \label{fig:competitors-quantitative}
\end{figure}
\addtolength{\tabcolsep}{4pt}

\subsection{Reenactment performance}
\label{sect:reenactment-performance}

\setlength{\widei}{0.13\textwidth}
\addtolength{\tabcolsep}{-5pt}
\begin{figure*}
    \centering
    \begin{tabular}{cccccc}
    \includegraphics[width=\widei]{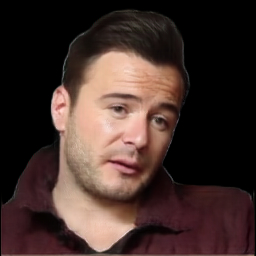}
    &\includegraphics[width=\widei]{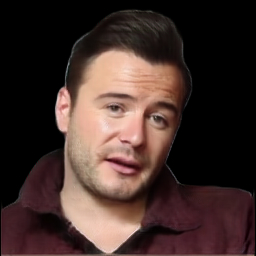}
    &\includegraphics[width=\widei]{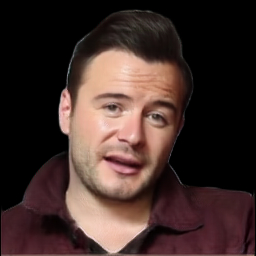}
    &\includegraphics[width=\widei]{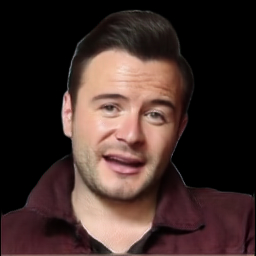}
    &\includegraphics[width=\widei]{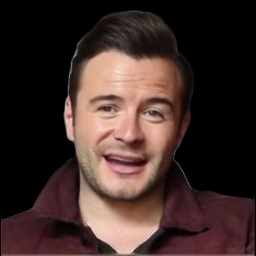}
    &\includegraphics[width=\widei]{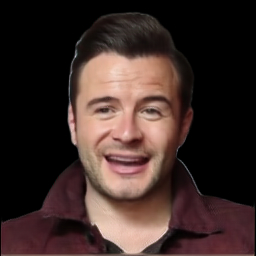}    
    \end{tabular}
    \begin{tabular}{cccccc}
    \includegraphics[width=\widei]{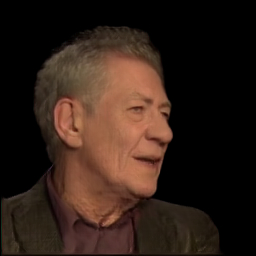}
    &\includegraphics[width=\widei]{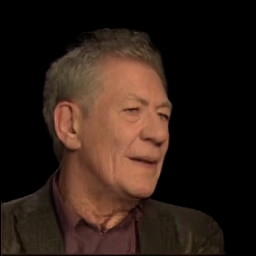}
    &\includegraphics[width=\widei]{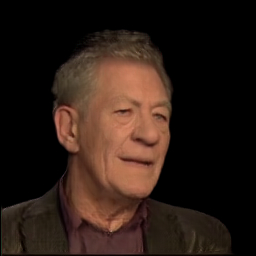}
    &\includegraphics[width=\widei]{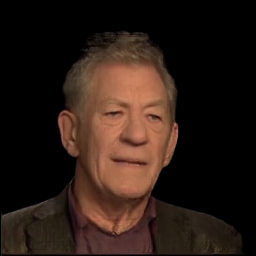}
    &\includegraphics[width=\widei]{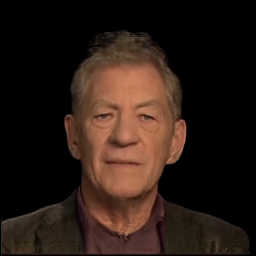}
    &\includegraphics[width=\widei]{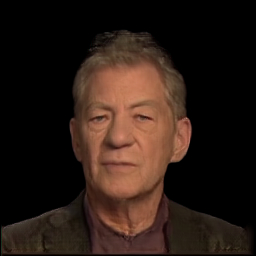}    
    \end{tabular}
    \begin{tabular}{cccccc}
    \includegraphics[width=\widei]{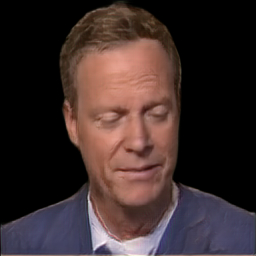}
    &\includegraphics[width=\widei]{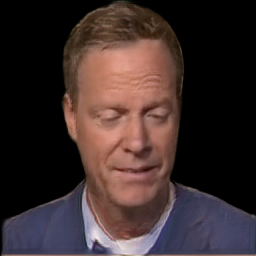}
    &\includegraphics[width=\widei]{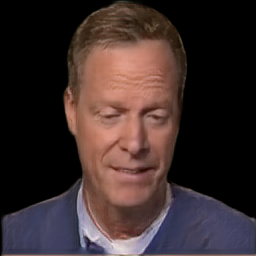}
    &\includegraphics[width=\widei]{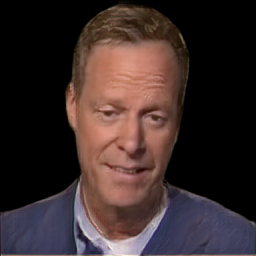}
    &\includegraphics[width=\widei]{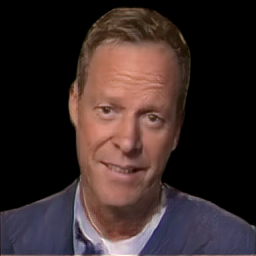}
    &\includegraphics[width=\widei]{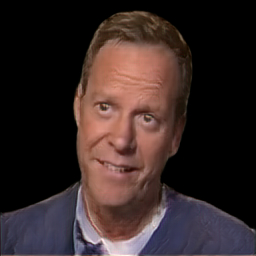}    
    \end{tabular}
    \caption{Reenactment by interpolation between two pose vectors across spherical trajectory in the pose descriptor space. Our system successfully creates visually smooth and identity preserving reenactment.}
    \label{fig:interpolation}
\end{figure*}
\addtolength{\tabcolsep}{5pt}

\paragraph{Quantitative evaluation.} We compare the performance of the seven reenactment systems listed above in the cross-person setting. To do this, we randomly choose 30 people from the test split of VoxCeleb2 and learn talking head models $T_1, \dots, T_{30}$ for them. Each model $T_k$ is created from 32 random frames of a video $I_1^k, \dots, I_{32}^k$. All models except \textsf{X2Face} are fine-tuned to those 32 frames for 600 optimization steps. Using these models, we compute two metrics per system, \textit{identity error} $\mathcal{I}_T$ and \textit{pose reconstruction error} $\mathcal{P}_T$.

The identity error $\mathcal{I}_T$ estimates how closely the resulting talking heads resemble the original person $k$ that the model was learned for. For that, we use the \textit{ArcFace}~\cite{Deng19} face recognition network $R$ that outputs identity descriptors (vectors). We compute the averaged reference descriptor $r_k = \frac{1}{32} \sum_{j=1}^{32} R(I_j^k)$ from the fine-tuning dataset $I_1^k, \dots, I_{32}^k$, and use the cosine similarity ($\text{csim}$) to compare it with the descriptors obtained from cross-person reenactment results. Cross-person reenactment is performed by driving $T_k$ with all other 29 people. To obtain the final error, we average (one minus) similarities over all 30 people in the test set. Formally,

$$
\mathcal{I}_T = \frac{1}{30 \cdot 29 \cdot 32} \sum_{k=1}^{30} \sum_{\substack{i=1 \\ i \neq k}}^{30} \sum_{j=1}^{32} \left[ 1 - \text{csim}\left( R\left( T_k(I^i_j) \right), ~r_k \right) \right].
$$

The \textit{pose reconstruction error} $\mathcal{P}_T$, on the other hand, is designed to quantify how well the system replays driver's pose and facial expression, and is defined in terms of facial landmarks. Since sets of landmarks can only be compared directly for the same person, we restrict the test dataset to self-reenactment pairs, i.e. we only drive $T_k$ with $I^k$. However, because $T_k$ has learned on $I_1^k, \dots, I_{32}^k$, we use \textbf{another} 32 hold-out frames from the same video $I_{33}^k, \dots, I_{64}^k$ to avoid overfitting. We employ an off-the-shelf 2D facial landmarks prediction algorithm~\cite{Bulat17} $L$ to obtain landmarks in both the driver $I_j^k$ and the reenactment result $T_k(I_j^k)$. In our case, a measure $d_{\text{landmarks}}(l_1,~l_2)$ of how close landmarks $l_2$ approximate reference landmarks $l_1$ is the average distance between corresponding landmarks normalized by inter-ocular distance. As before, we compute $d$ for all drivers and average across all 30 people:

$$
\mathcal{P}_T = \frac{1}{30 \cdot 32} \sum_{k=1}^{30} \sum_{j=33}^{64} d_{\text{landmarks}} \left( L\left( I_j^k \right),~ L\left( T_k(I_j^k) \right) \right).
$$

The plot in \fig{competitors-quantitative} has these two metrics evaluated for the compared models. A perfect system $T$ would have $\mathcal{I}_T = \mathcal{P}_T = 0$, i.e. the closer to the lower left corner, the better. In these terms, our full model is strictly better than all systems except \textsf{FSTH}, which is slightly better in one of the metrics but much worse in the other, and which benefits from an external keypoint detector.

\setlength{\oldtabcolsep}{\tabcolsep}
\setlength{\tabcolsep}{1pt}
\setlength{\widei}{0.123\textwidth}
\begin{figure*}
    \centering
    \begin{tabular}{M{\widei}M{\widei}M{\widei}M{\widei}M{\widei}M{\widei}M{\widei}M{\widei}}
    \includegraphics[clip,width=\widei]{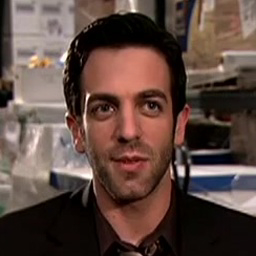} &
    \includegraphics[clip,trim=0 0 0 0,width=\widei]{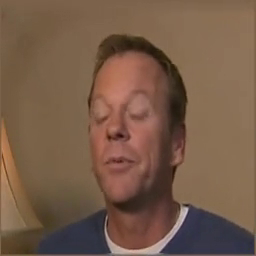} &
    \includegraphics[clip,trim=0 0 0 0,width=\widei]{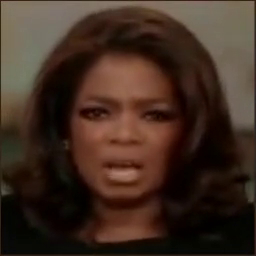} &
    \includegraphics[clip,trim=0 0 0 0,width=\widei]{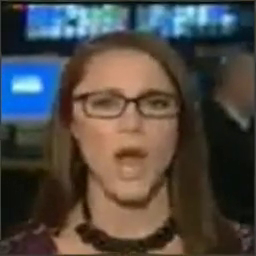} &
    \includegraphics[clip,trim=0 0 0 0,width=\widei]{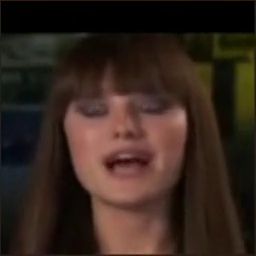} &
    \includegraphics[clip,trim=0 0 0 0,width=\widei]{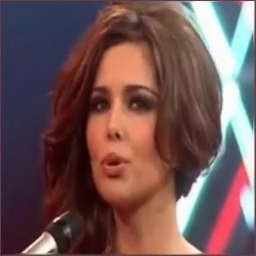} &
    \includegraphics[clip,trim=0 0 0 0,width=\widei]{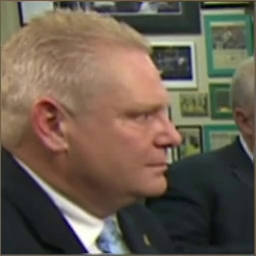} &
    \includegraphics[clip,trim=0 0 0 0,width=\widei]{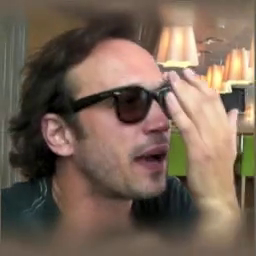} \\
    \textbf{\textsf{FSTH}} &
    \includegraphics[clip,trim=0 0 0 0,width=\widei]{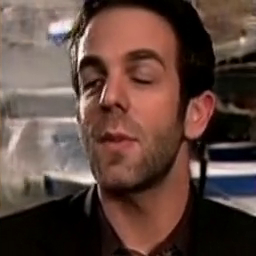} &
    \includegraphics[clip,trim=0 0 0 0,width=\widei]{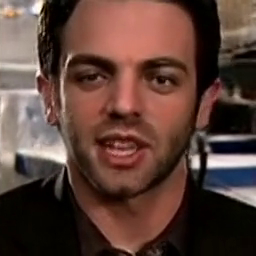} &
    \includegraphics[clip,trim=0 0 0 0,width=\widei]{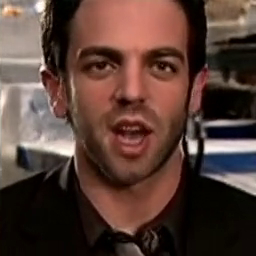} &
    \includegraphics[clip,trim=0 0 0 0,width=\widei]{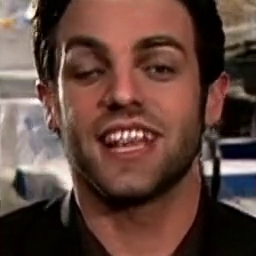} &
    \includegraphics[clip,trim=0 0 0 0,width=\widei]{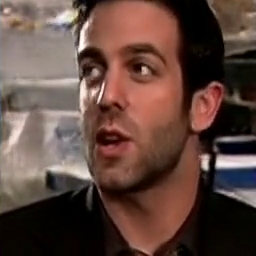} &
    \includegraphics[clip,trim=0 0 0 0,width=\widei]{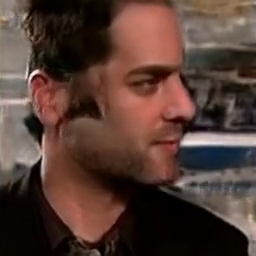} &
    \includegraphics[clip,trim=0 0 0 0,width=\widei]{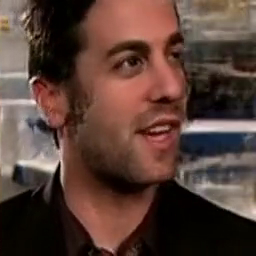} \\
    \textbf{\textsf{FSTH+}} &
    \includegraphics[clip,trim=0 0 0 0,width=\widei]{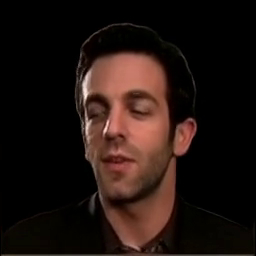} &
    \includegraphics[clip,trim=0 0 0 0,width=\widei]{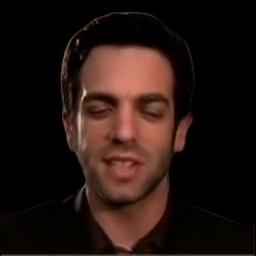} &
    \includegraphics[clip,trim=0 0 0 0,width=\widei]{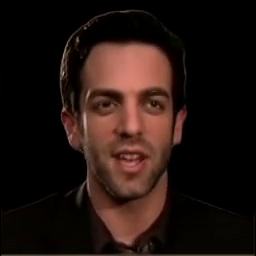} &
    \includegraphics[clip,trim=0 0 0 0,width=\widei]{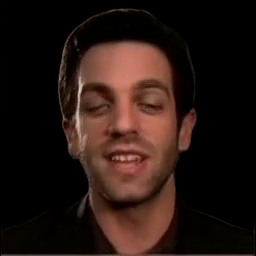} &
    \includegraphics[clip,trim=0 0 0 0,width=\widei]{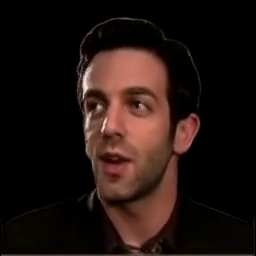} &
    \includegraphics[clip,trim=0 0 0 0,width=\widei]{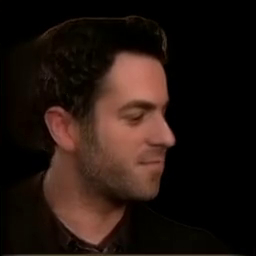} &
    \includegraphics[clip,trim=0 0 0 0,width=\widei]{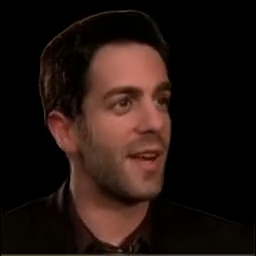} \\
    \textbf{\textsf{X2Face}} &
    \includegraphics[clip,trim=0 0 0 0,width=\widei]{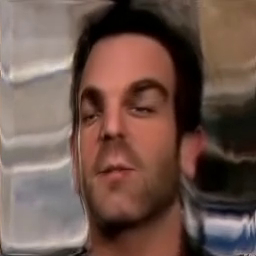} &
    \includegraphics[clip,trim=0 0 0 0,width=\widei]{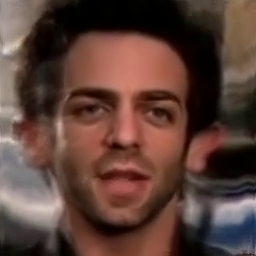} &
    \includegraphics[clip,trim=0 0 0 0,width=\widei]{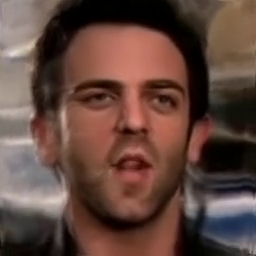} &
    \includegraphics[clip,trim=0 0 0 0,width=\widei]{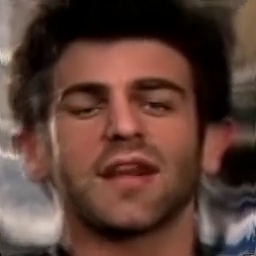} &
    \includegraphics[clip,trim=0 0 0 0,width=\widei]{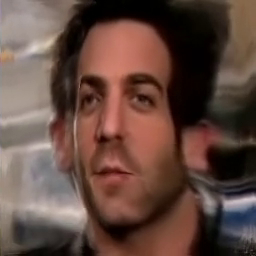} &
    \includegraphics[clip,trim=0 0 0 0,width=\widei]{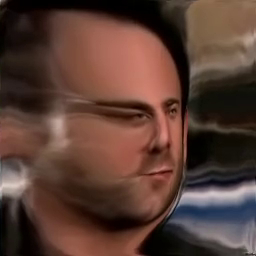} &
    \includegraphics[clip,trim=0 0 0 0,width=\widei]{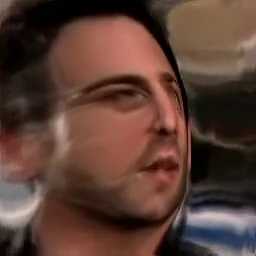} \\
    \textbf{\textsf{X2Face+}} &
    \includegraphics[clip,trim=0 0 0 0,width=\widei]{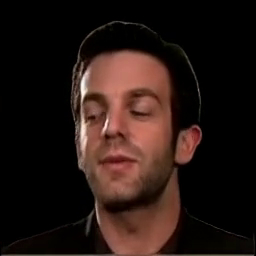} &
    \includegraphics[clip,trim=0 0 0 0,width=\widei]{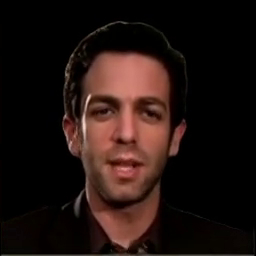} &
    \includegraphics[clip,trim=0 0 0 0,width=\widei]{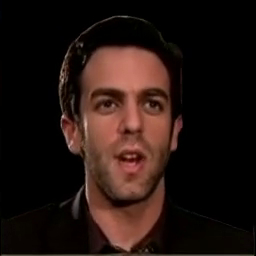} &
    \includegraphics[clip,trim=0 0 0 0,width=\widei]{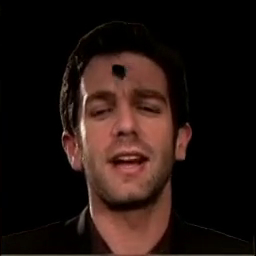} &
    \includegraphics[clip,trim=0 0 0 0,width=\widei]{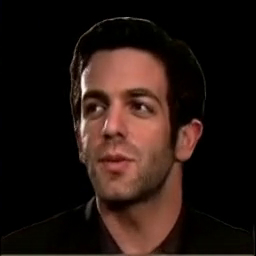} &
    \includegraphics[clip,trim=0 0 0 0,width=\widei]{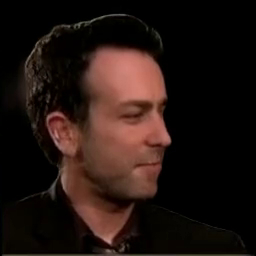} &
    \includegraphics[clip,trim=0 0 0 0,width=\widei]{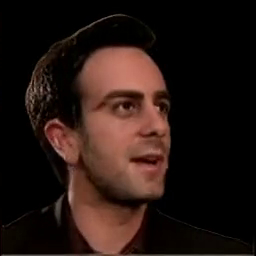} \\
    \textbf{\textsf{FAb-Net+}} &
    \includegraphics[clip,trim=0 0 0 0,width=\widei]{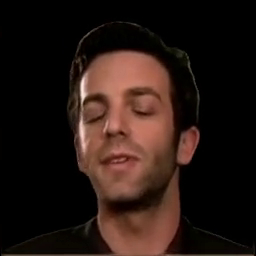} &
    \includegraphics[clip,trim=0 0 0 0,width=\widei]{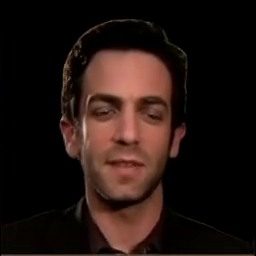} &
    \includegraphics[clip,trim=0 0 0 0,width=\widei]{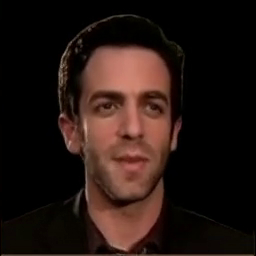} &
    \includegraphics[clip,trim=0 0 0 0,width=\widei]{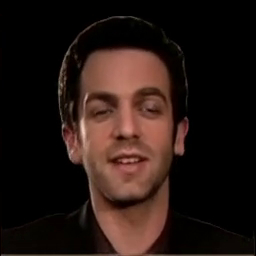} &
    \includegraphics[clip,trim=0 0 0 0,width=\widei]{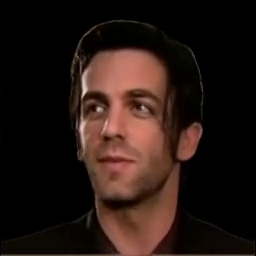} &
    \includegraphics[clip,trim=0 0 0 0,width=\widei]{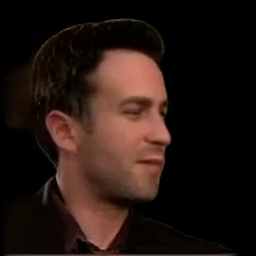} &
    \includegraphics[clip,trim=0 0 0 0,width=\widei]{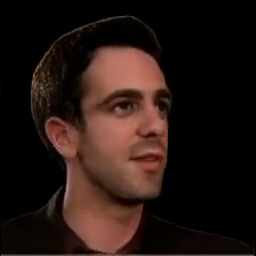} \\
    \textbf{\textsf{3DMM+}} &
    \includegraphics[clip,trim=0 0 0 0,width=\widei]{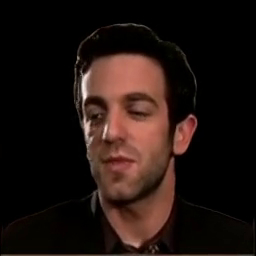} &
    \includegraphics[clip,trim=0 0 0 0,width=\widei]{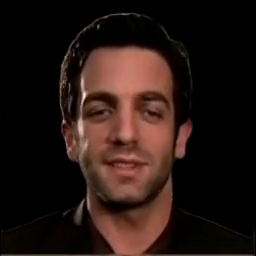} &
    \includegraphics[clip,trim=0 0 0 0,width=\widei]{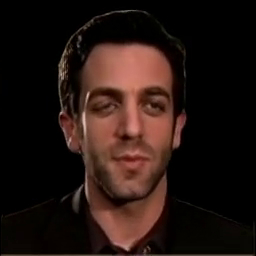} &
    \includegraphics[clip,trim=0 0 0 0,width=\widei]{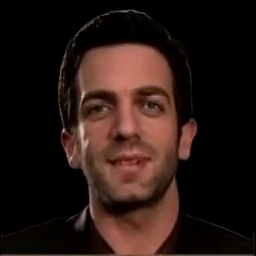} &
    \includegraphics[clip,trim=0 0 0 0,width=\widei]{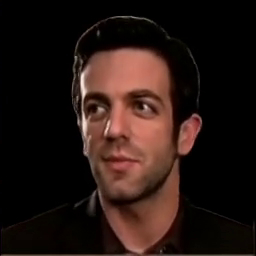} &
    \includegraphics[clip,trim=0 0 0 0,width=\widei]{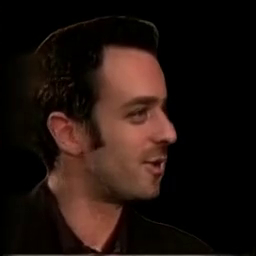} &
    \includegraphics[clip,trim=0 0 0 0,width=\widei]{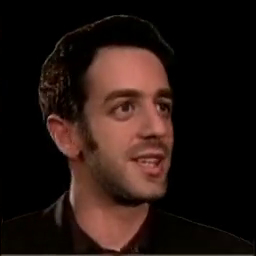} \\
    \textbf{\textsf{Ours}} &
    \includegraphics[clip,trim=0 0 0 0,width=\widei]{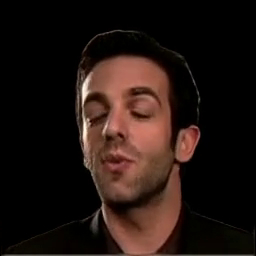} &
    \includegraphics[clip,trim=0 0 0 0,width=\widei]{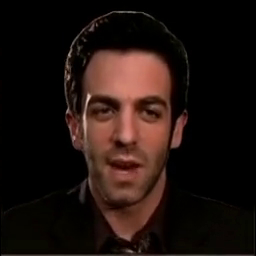} &
    \includegraphics[clip,trim=0 0 0 0,width=\widei]{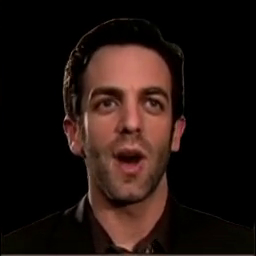} &
    \includegraphics[clip,trim=0 0 0 0,width=\widei]{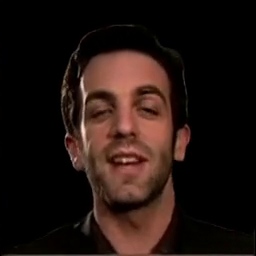} &
    \includegraphics[clip,trim=0 0 0 0,width=\widei]{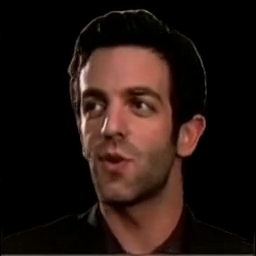} &
    \includegraphics[clip,trim=0 0 0 0,width=\widei]{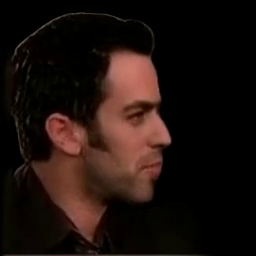} &
    \includegraphics[clip,trim=0 0 0 0,width=\widei]{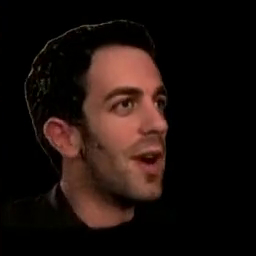} \\
    \end{tabular}
    \caption{Comparison of cross-person reenactment for several systems on VoxCeleb2 test set. The top left image is one of the 32 identity source frames. The other images in the top row are pose drivers. Our method better preserves the identity of the target person and successfully transfers the mimics from the driver person.}
    \label{fig:competitors}
\end{figure*}
\setlength{\tabcolsep}{\oldtabcolsep}

\paragraph{Qualitative comparison.} \fig{competitors} gives a qualitative comparison of the reenactment systems described above. It is evident that \textsf{FSTH}, being driven by rasterized landmarks, relies heavily on the driver's facial proportions, and thus is not person-agnostic. Its modified version \textsf{FSTH+} does a better job having more representational power around vectorized keypoints; still, there are visible "identity bleeding" (e.g. compare head width in columns 1 and 2) and errors in prominent facial expressions, such as closing eyes. The warping-based method \textsf{X2Face} fails on slight rotations already.

Two similar methods, \textsf{X2Face+} and \textsf{FAb-Net+}, both provide strong baselines despite some signs of identity mismatch, for example, traces of eyeglasses in column 7 and long hair seeping in from the pose driver in column 5. It is important to note that although pose descriptors from those methods are not person-agnostic, we still apply pose augmentations during training. In the ablation study below, we demonstrate that cross-person reenactment performance drops dramatically when we remove pose augmentations in these two methods.

The \textsf{3DMM+} method has a very tight bottleneck of interpretable parameters, and therefore its identity gap is very small. However, apparently for the same reason, it is not as good at rendering correct subtle facial expressions. Our full system is able to accurately represent pose driver's facial expression while preserving identity of the target person. 

In addition, we also show reenactment by interpolation in the pose space for our system in \fig{interpolation}, which demonstrates smooth pose changes.

\paragraph{Temporal smoothness.} The supplementary video demonstrates the capability of our descriptor to create temporally smooth reenactment without any temporal smoothing of the extracted pose (provided that the results of bounding box detection are temporally smooth). At the same time, we have found that achieving temporally smooth reenactment with keypoint-driven systems (\textsf{FSTH}, \textsf{FSTH+}) requires a lot of keypoint smoothing.

\section{Discussion}

We have presented and evaluated a neural head reenactment that uses latent pose descriptors and is able to achieve realistic reenactment. Unlike the predecessor system \cite{Zakharov19} that used the keypoints as pose descriptor, our system uses the pose descriptors without explicit supervision purely based on the reconstruction losses. The only weak form of supervision comes from the segmentation masks. Our learned head pose descriptors outperform previous unsupervised descriptors at the tasks of pose-based retrieval, as well as cross-person reenactment.

Our main, perhaps surprising, finding is that limiting capacity of the pose extraction network in our scheme is sufficient for pose/identity disentanglement. At the same time, it might happen that appropriate use of cyclic and/or adversarial losses may improve disentanglement even better. Perhaps because of the constraint on the network capacity, our pose descriptors and reenactment system has problems with capturing some subtle mimics, especially gaze direction (though it still does a better job than keypoint descriptors that lack gaze representation altogether). Another obvious avenue of research is learning pose descriptor and the entire system in a semi-supervised way.


\FloatBarrier
\ifnum\value{page}>8 \errmessage{Number of pages exceeded!!!!}\fi

{\small
\bibliographystyle{ieee}
\bibliography{main}

\begin{thebibliography}{10}\itemsep=-1pt

\bibitem{Blanz99}
V.~Blanz, T.~Vetter, et~al.
\newblock A morphable model for the synthesis of 3d faces.
\newblock In {\em Proc. {SIGGRAPH}}, volume~99, pages 187--194, 1999.

\bibitem{Bulat17}
A.~Bulat and G.~Tzimiropoulos.
\newblock How far are we from solving the 2d {\&} 3d face alignment problem?
  (and a dataset of 230, 000 3d facial landmarks).
\newblock In {\em Proc. {ICCV}}, pages 1021--1030, 2017.

\bibitem{Chang18}
F.-J. Chang, A.~T. Tran, T.~Hassner, I.~Masi, R.~Nevatia, and G.~Medioni.
\newblock Expnet: Landmark-free, deep, 3d facial expressions.
\newblock In {\em In proc. FG}, pages 122--129. IEEE, 2018.

\bibitem{Chung18}
J.~S. Chung, A.~Nagrani, and A.~Zisserman.
\newblock Voxceleb2: Deep speaker recognition.
\newblock In {\em INTERSPEECH}, 2018.

\bibitem{Chung17}
J.~S. Chung, A.~Senior, O.~Vinyals, and A.~Zisserman.
\newblock Lip reading sentences in the wild.
\newblock In {\em Proc. {CVPR}}, pages 3444--3453. IEEE, 2017.

\bibitem{Cootes01}
T.~F. Cootes, G.~J. Edwards, and C.~J. Taylor.
\newblock Active appearance models.
\newblock {\em {T-PAMI}}, (6):681--685, 2001.

\bibitem{Deng19}
J.~Deng, J.~Guo, X.~Niannan, and S.~Zafeiriou.
\newblock Arcface: Additive angular margin loss for deep face recognition.
\newblock In {\em CVPR}, 2019.

\bibitem{Denton17}
E.~L. Denton and V.~Birodkar.
\newblock Unsupervised learning of disentangled representations from video.
\newblock In {\em Proc. {NeurIPS}}, pages 4414--4423, 2017.

\bibitem{Ekman77}
P.~Ekman and W.~V. Friesen.
\newblock Facial action coding system.
\newblock 1977.

\bibitem{Fu19}
C.~Fu, Y.~Hu, X.~Wu, G.~Wang, Q.~Zhang, and R.~He.
\newblock High fidelity face manipulation with extreme pose and expression.
\newblock {\em arXiv preprint arXiv:1903.12003}, 2019.

\bibitem{Gong19}
K.~Gong, Y.~Gao, X.~Liang, X.~Shen, M.~Wang, and L.~Lin.
\newblock Graphonomy: Universal human parsing via graph transfer learning.
\newblock In {\em CVPR}, 2019.

\bibitem{Goodfellow14}
I.~Goodfellow, J.~Pouget-Abadie, M.~Mirza, B.~Xu, D.~Warde-Farley, S.~Ozair,
  A.~Courville, and Y.~Bengio.
\newblock Generative adversarial nets.
\newblock In {\em Proc. {NIPS}}, 2014.

\bibitem{gross2008}
R.~Gross, I.~Matthews, J.~Cohn, T.~Kanade, and S.~Baker.
\newblock Multi-pie.
\newblock In {\em Proceedings of the IEEE International Conference on Automatic
  Face and Gesture Recognition}. IEEE Computer Society, September 2008.

\bibitem{Guler18}
R.~A. G{\"u}ler, N.~Neverova, and I.~Kokkinos.
\newblock {DensePose}: Dense human pose estimation in the wild.
\newblock In {\em Proc. {CVPR}}, June 2018.

\bibitem{Guler17}
R.~A. G{\"u}ler, G.~Trigeorgis, E.~Antonakos, P.~Snape, S.~Zafeiriou, and
  I.~Kokkinos.
\newblock {DenseReg}: Fully convolutional dense shape regression in-the-wild.
\newblock In {\em CVPR}, volume~2, page~5, 2017.

\bibitem{Huang17}
X.~Huang and S.~Belongie.
\newblock Arbitrary style transfer in real-time with adaptive instance
  normalization.
\newblock In {\em Proc. {ICCV}}, 2017.

\bibitem{Huang18}
X.~Huang, M.-Y. Liu, S.~Belongie, and J.~Kautz.
\newblock Multimodal unsupervised image-to-image translation.
\newblock In {\em Proc. {ECCV}}, 2018.

\bibitem{Ioffe15}
S.~Ioffe and C.~Szegedy.
\newblock Batch normalization: Accelerating deep network training by reducing
  internal covariate shift.
\newblock In {\em Proc. {ICML}}, ICML'15, pages 448--456, 2015.

\bibitem{Jakab18}
T.~Jakab, A.~Gupta, H.~Bilen, and A.~Vedaldi.
\newblock Unsupervised learning of object landmarks through conditional image
  generation.
\newblock In {\em Proc. {NeurIPS}}, pages 4016--4027, 2018.

\bibitem{Karras19}
T.~Karras, S.~Laine, and T.~Aila.
\newblock A style-based generator architecture for generative adversarial
  networks.
\newblock In {\em Proc. {CVPR}}, June 2019.

\bibitem{Kazemi14}
V.~Kazemi and J.~Sullivan.
\newblock One millisecond face alignment with an ensemble of regression trees.
\newblock {\em Proc. {CVPR}}, pages 1867--1874, 2014.

\bibitem{Kim18}
H.~Kim, P.~Garrido, A.~Tewari, W.~Xu, J.~Thies, M.~Nie{\ss}ner, P.~P{\'e}rez,
  C.~Richardt, M.~Zollh{\"o}fer, and C.~Theobalt.
\newblock Deep video portraits.
\newblock In {\em Proc. {SIGGRAPH}}, 2018.

\bibitem{Kim18b}
H.~Kim and A.~Mnih.
\newblock Disentangling by factorising.
\newblock In {\em Proc. {ICML}}, pages 2654--2663, 2018.

\bibitem{Liu19}
M.~Liu, X.~Huang, A.~Mallya, T.~Karras, T.~Aila, J.~Lehtinen, and J.~Kautz.
\newblock Few-shot unsupervised image-to-image translation.
\newblock In {\em Proc. {ICCV}}, 2019.

\bibitem{Liu2015}
Z.~Liu, P.~Luo, X.~Wang, and X.~Tang.
\newblock Deep learning face attributes in the wild.
\newblock In {\em Proc. {ICCV}}, December 2015.

\bibitem{Lombardi18}
S.~Lombardi, J.~Saragih, T.~Simon, and Y.~Sheikh.
\newblock Deep appearance models for face rendering.
\newblock {\em ACM Transactions on Graphics (TOG)}, 37(4):68, 2018.

\bibitem{Milletari16}
F.~Milletari, N.~Navab, and S.-A. Ahmadi.
\newblock V-net: Fully convolutional neural networks for volumetric medical
  image segmentation.
\newblock pages 565--571, 10 2016.

\bibitem{Miyato18b}
T.~Miyato, T.~Kataoka, M.~Koyama, and Y.~Yoshida.
\newblock Spectral normalization for generative adversarial networks.
\newblock In {\em International Conference on Learning Representations}, 2018.

\bibitem{Nagrani17}
A.~Nagrani, J.~S. Chung, and A.~Zisserman.
\newblock Voxceleb: a large-scale speaker identification dataset.
\newblock In {\em INTERSPEECH}, 2017.

\bibitem{Pumarola18}
A.~Pumarola, A.~Agudo, A.~M. Martinez, A.~Sanfeliu, and F.~Moreno-Noguer.
\newblock Ganimation: Anatomically-aware facial animation from a single image.
\newblock In {\em Proc. {ECCV}}, pages 818--833, 2018.

\bibitem{Sandler18}
M.~Sandler, A.~Howard, M.~Zhu, A.~Zhmoginov, and L.-C. Chen.
\newblock Mobilenetv2: Inverted residuals and linear bottlenecks.
\newblock In {\em Proc. {CVPR}}, June 2018.

\bibitem{Siarohin19}
A.~Siarohin, S.~Lathuili{\`{e}}re, S.~Tulyakov, E.~Ricci, and N.~Sebe.
\newblock First order motion model for image animation.
\newblock In {\em Proc. {NeurIPS}}, pages 7135--7145, 2019.

\bibitem{Suwajanakorn17}
S.~Suwajanakorn, S.~M. Seitz, and I.~Kemelmacher-Shlizerman.
\newblock Synthesizing {O}bama: learning lip sync from audio.
\newblock {\em ACM Transactions on Graphics (TOG)}, 36(4):95, 2017.

\bibitem{Thewlis19}
J.~Thewlis, S.~Albanie, H.~Bilen, and A.~Vedaldi.
\newblock Unsupervised learning of landmarks by descriptor vector exchange.
\newblock In {\em Proc. {ICCV}}, 2019.

\bibitem{Tripathy19}
S.~Tripathy, J.~Kannala, and E.~Rahtu.
\newblock Icface: Interpretable and controllable face reenactment using gans.
\newblock {\em CoRR}, abs/1904.01909, 2019.

\bibitem{Wang19}
T.~Wang, M.~Liu, A.~Tao, G.~Liu, J.~Kautz, and B.~Catanzaro.
\newblock Few-shot video-to-video synthesis.
\newblock {\em CoRR}, abs/1910.12713, 2019.

\bibitem{Wang18c}
T.-C. Wang, M.-Y. Liu, J.-Y. Zhu, G.~Liu, A.~Tao, J.~Kautz, and B.~Catanzaro.
\newblock Video-to-video synthesis.
\newblock {\em Proc. {NeurIPS}}, 2018.

\bibitem{Wiles18b}
O.~Wiles, A.~S. Koepke, and A.~Zisserman.
\newblock Self-supervised learning of a facial attribute embedding from video.
\newblock In {\em Proc. {BMVC}}, 2018.

\bibitem{Wiles18a}
O.~Wiles, A.~S. Koepke, and A.~Zisserman.
\newblock X2face: A network for controlling face generation using images,
  audio, and pose codes.
\newblock In {\em Proc. {ECCV}}, September 2018.

\bibitem{Xiao19}
F.~Xiao, H.~Liu, and Y.~J. Lee.
\newblock Identity from here, pose from there: Self-supervised disentanglement
  and generation of objects using unlabeled videos.
\newblock In {\em Proc. {ICCV}}, October 2019.

\bibitem{Xie17}
S.~Xie, R.~Girshick, P.~Dollar, Z.~Tu, and K.~He.
\newblock Aggregated residual transformations for deep neural networks.
\newblock In {\em Proc. {CVPR}}, July 2017.

\bibitem{Zakharov19}
E.~Zakharov, A.~Shysheya, E.~Burkov, and V.~Lempitsky.
\newblock Few-shot adversarial learning of realistic neural talking head
  models.
\newblock In {\em Proc. {ICCV}}, October 2019.

\bibitem{Zhang17}
S.~Zhang, X.~Zhu, Z.~Lei, H.~Shi, X.~Wang, and S.~Z. Li.
\newblock S3fd: Single shot scale-invariant face detector.
\newblock In {\em Proc. {ICCV}}, Oct 2017.

\bibitem{Zhang18}
Y.~Zhang, Y.~Guo, Y.~Jin, Y.~Luo, Z.~He, and H.~Lee.
\newblock Unsupervised discovery of object landmarks as structural
  representations.
\newblock In {\em Proc. {CVPR}}, pages 2694--2703, 2018.

\bibitem{Zhu17}
J.-Y. Zhu, T.~Park, P.~Isola, and A.~A. Efros.
\newblock Unpaired image-to-image translation using cycle-consistent
  adversarial networks.
\newblock In {\em Proc. {ICCV}}, 2017.

\end{thebibliography}
}

\clearpage
\maketitle

\appendix
\section{Supplementary material}

\subsection{Reenactment performance}

In \fig{competitors2}, we provide additional qualitative comparison of talking head systems listed in \sect{compared-methods} in a cross-person reenactment scenario.

\setlength{\oldtabcolsep}{\tabcolsep}
\setlength{\tabcolsep}{1pt}
\setlength{\widei}{0.0984\textwidth}
\begin{figure*}[!ht]
    \centering
    \begin{tabular}{M{\widei}M{\widei}M{\widei}M{\widei}M{\widei}M{\widei}M{\widei}M{\widei}M{\widei}M{\widei}}
    \includegraphics[clip,width=\widei]{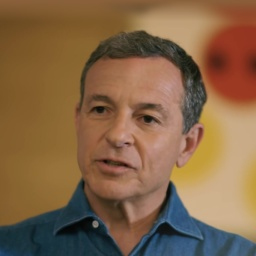} &
    \includegraphics[clip,trim=0 0 0 0,width=\widei]{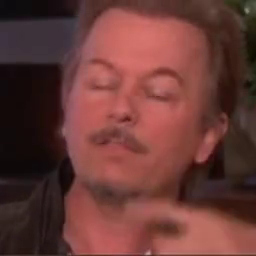} &
    \includegraphics[clip,trim=0 0 0 0,width=\widei]{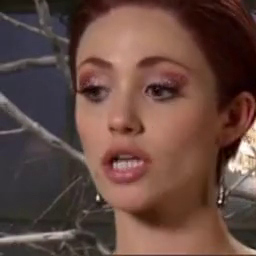} &
    \includegraphics[clip,trim=0 0 0 0,width=\widei]{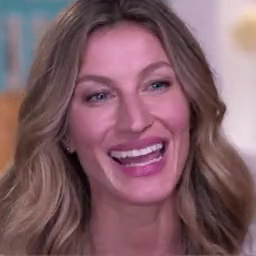} &
    \includegraphics[clip,trim=0 0 0 0,width=\widei]{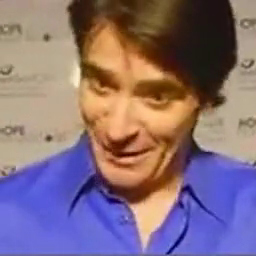} &
    \includegraphics[clip,trim=0 0 0 0,width=\widei]{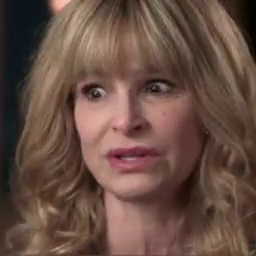} &
    \includegraphics[clip,trim=0 0 0 0,width=\widei]{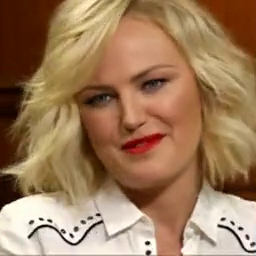} &
    \includegraphics[clip,trim=0 0 0 0,width=\widei]{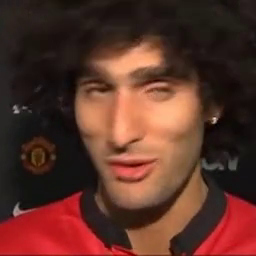} &
    \includegraphics[clip,trim=0 0 0 0,width=\widei]{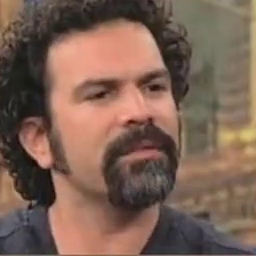} &
    \includegraphics[clip,trim=0 0 0 0,width=\widei]{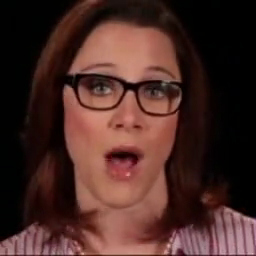} \\
    \textbf{\textsf{FSTH}} &
    \includegraphics[clip,trim=0 0 0 0,width=\widei]{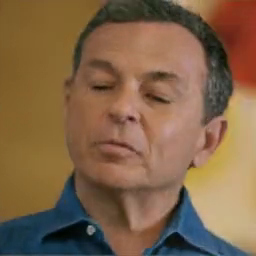} &
    \includegraphics[clip,trim=0 0 0 0,width=\widei]{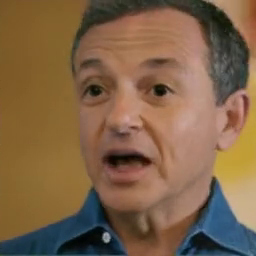} &
    \includegraphics[clip,trim=0 0 0 0,width=\widei]{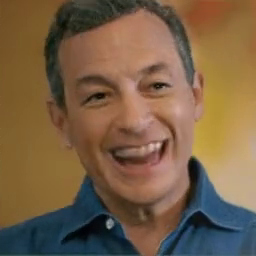} &
    \includegraphics[clip,trim=0 0 0 0,width=\widei]{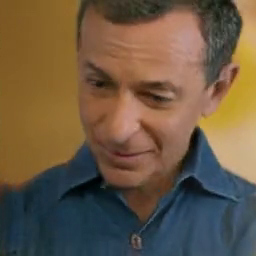} &
    \includegraphics[clip,trim=0 0 0 0,width=\widei]{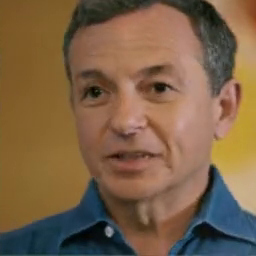} &
    \includegraphics[clip,trim=0 0 0 0,width=\widei]{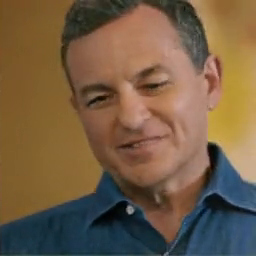} &
    \includegraphics[clip,trim=0 0 0 0,width=\widei]{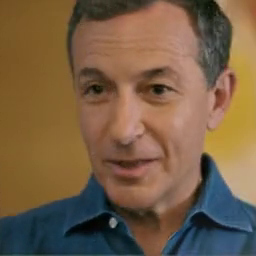} &
    \includegraphics[clip,trim=0 0 0 0,width=\widei]{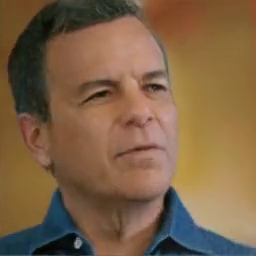} &
    \includegraphics[clip,trim=0 0 0 0,width=\widei]{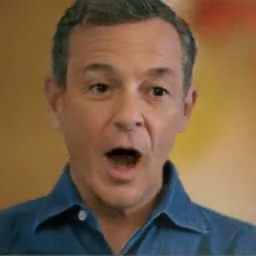} \\
    \begin{tabular}{c}\textbf{\textsf{FSTH+}}\end{tabular} &
    \includegraphics[clip,trim=0 0 0 0,width=\widei]{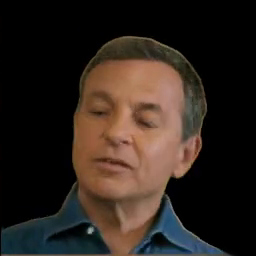} &
    \includegraphics[clip,trim=0 0 0 0,width=\widei]{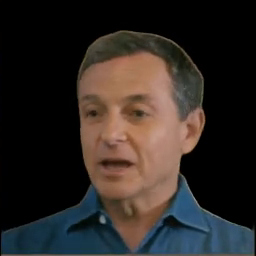} &
    \includegraphics[clip,trim=0 0 0 0,width=\widei]{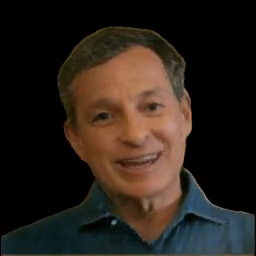} &
    \includegraphics[clip,trim=0 0 0 0,width=\widei]{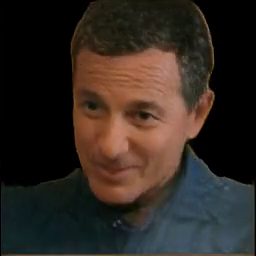} &
    \includegraphics[clip,trim=0 0 0 0,width=\widei]{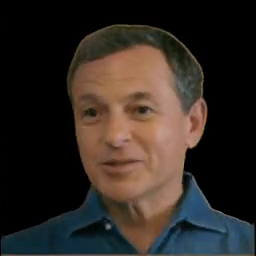} &
    \includegraphics[clip,trim=0 0 0 0,width=\widei]{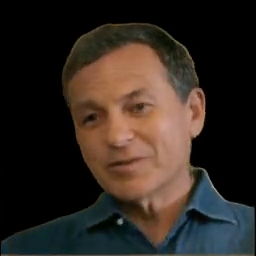} &
    \includegraphics[clip,trim=0 0 0 0,width=\widei]{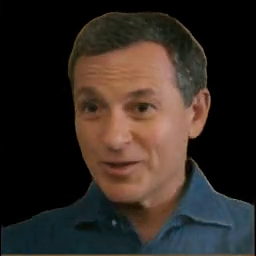} &
    \includegraphics[clip,trim=0 0 0 0,width=\widei]{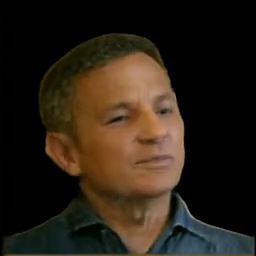} &
    \includegraphics[clip,trim=0 0 0 0,width=\widei]{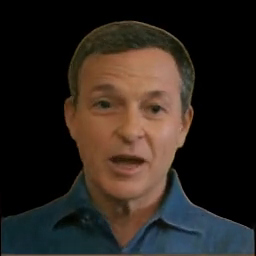} \\
    \begin{tabular}{c}\textbf{\textsf{X2Face}}\end{tabular} &
    \includegraphics[clip,trim=0 0 0 0,width=\widei]{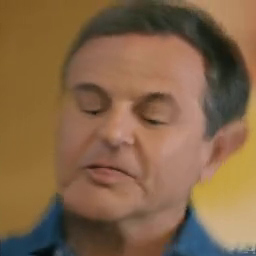} &
    \includegraphics[clip,trim=0 0 0 0,width=\widei]{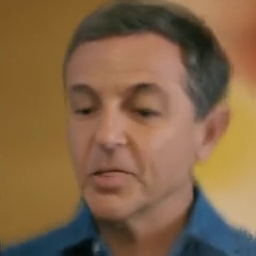} &
    \includegraphics[clip,trim=0 0 0 0,width=\widei]{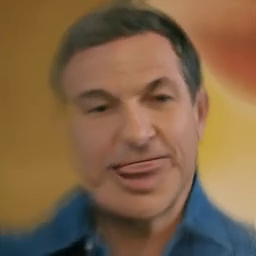} &
    \includegraphics[clip,trim=0 0 0 0,width=\widei]{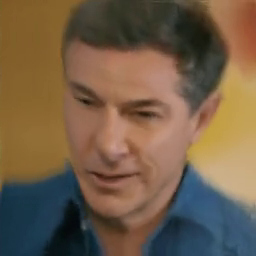} &
    \includegraphics[clip,trim=0 0 0 0,width=\widei]{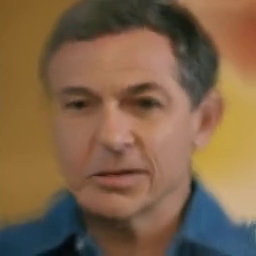} &
    \includegraphics[clip,trim=0 0 0 0,width=\widei]{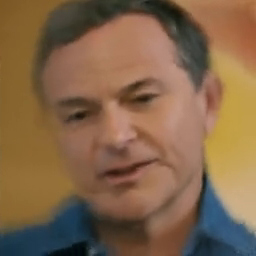} &
    \includegraphics[clip,trim=0 0 0 0,width=\widei]{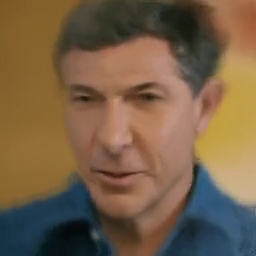} &
    \includegraphics[clip,trim=0 0 0 0,width=\widei]{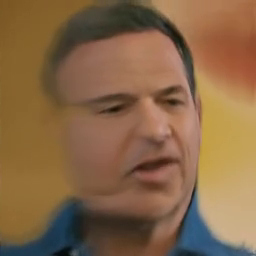} &
    \includegraphics[clip,trim=0 0 0 0,width=\widei]{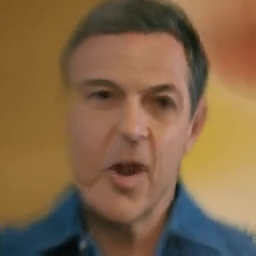} \\
    \begin{tabular}{c}\textbf{\textsf{X2Face+}}\end{tabular} &
    \includegraphics[clip,trim=0 0 0 0,width=\widei]{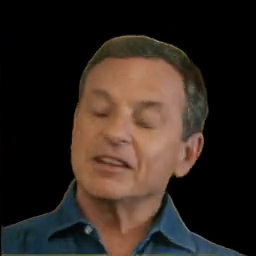} &
    \includegraphics[clip,trim=0 0 0 0,width=\widei]{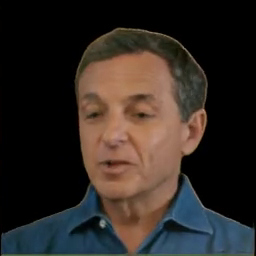} &
    \includegraphics[clip,trim=0 0 0 0,width=\widei]{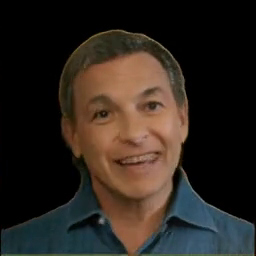} &
    \includegraphics[clip,trim=0 0 0 0,width=\widei]{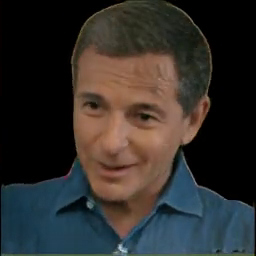} &
    \includegraphics[clip,trim=0 0 0 0,width=\widei]{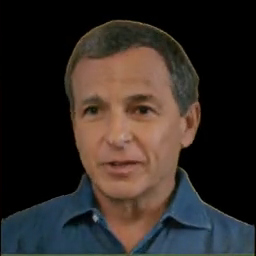} &
    \includegraphics[clip,trim=0 0 0 0,width=\widei]{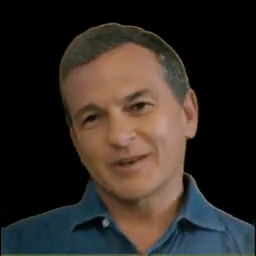} &
    \includegraphics[clip,trim=0 0 0 0,width=\widei]{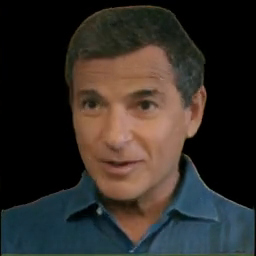} &
    \includegraphics[clip,trim=0 0 0 0,width=\widei]{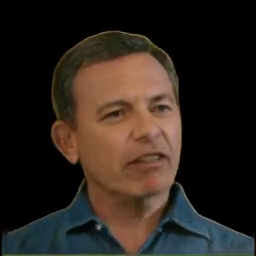} &
    \includegraphics[clip,trim=0 0 0 0,width=\widei]{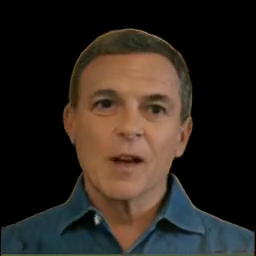} \\
    \begin{tabular}{c}\textbf{\textsf{FAb-Net+}}\end{tabular} &
    \includegraphics[clip,trim=0 0 0 0,width=\widei]{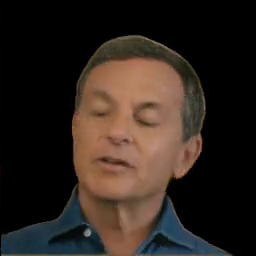} &
    \includegraphics[clip,trim=0 0 0 0,width=\widei]{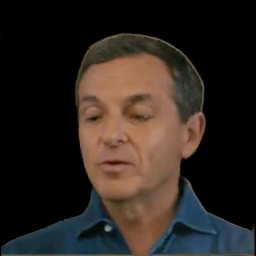} &
    \includegraphics[clip,trim=0 0 0 0,width=\widei]{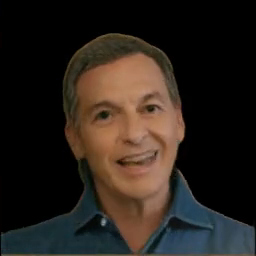} &
    \includegraphics[clip,trim=0 0 0 0,width=\widei]{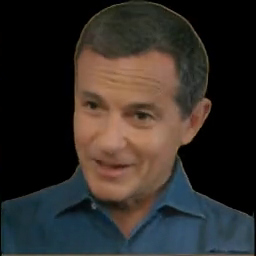} &
    \includegraphics[clip,trim=0 0 0 0,width=\widei]{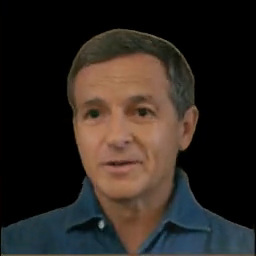} &
    \includegraphics[clip,trim=0 0 0 0,width=\widei]{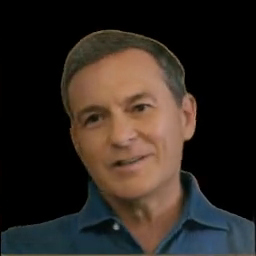} &
    \includegraphics[clip,trim=0 0 0 0,width=\widei]{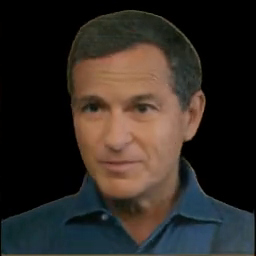} &
    \includegraphics[clip,trim=0 0 0 0,width=\widei]{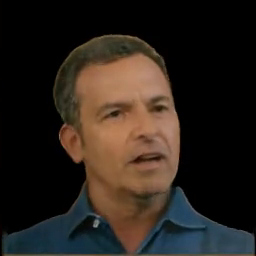} &
    \includegraphics[clip,trim=0 0 0 0,width=\widei]{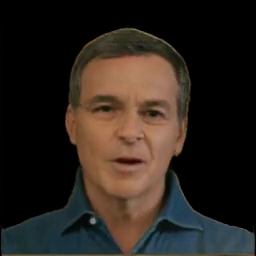} \\
    \begin{tabular}{c}\textbf{\textsf{3DMM+}}\end{tabular} &
    \includegraphics[clip,trim=0 0 0 0,width=\widei]{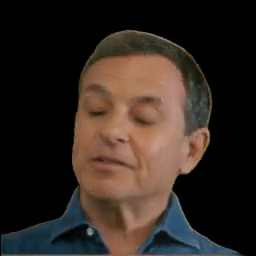} &
    \includegraphics[clip,trim=0 0 0 0,width=\widei]{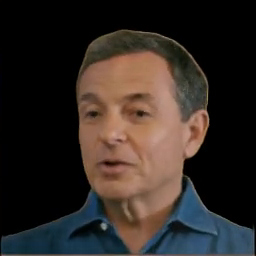} &
    \includegraphics[clip,trim=0 0 0 0,width=\widei]{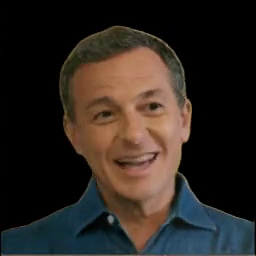} &
    \includegraphics[clip,trim=0 0 0 0,width=\widei]{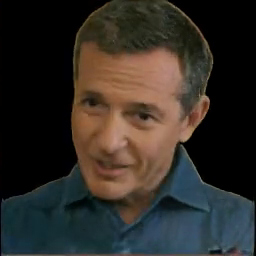} &
    \includegraphics[clip,trim=0 0 0 0,width=\widei]{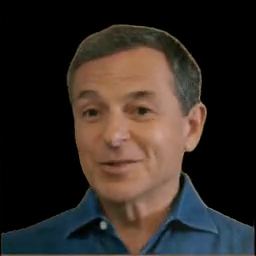} &
    \includegraphics[clip,trim=0 0 0 0,width=\widei]{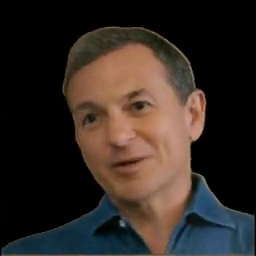} &
    \includegraphics[clip,trim=0 0 0 0,width=\widei]{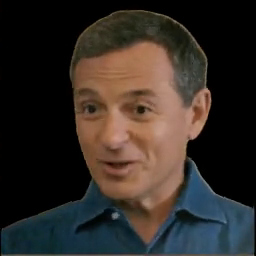} &
    \includegraphics[clip,trim=0 0 0 0,width=\widei]{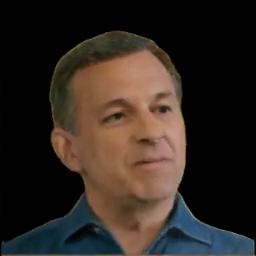} &
    \includegraphics[clip,trim=0 0 0 0,width=\widei]{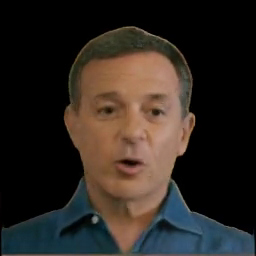} \\
    \begin{tabular}{c}\textbf{\textsf{Ours}}\end{tabular} &
    \includegraphics[clip,trim=0 0 0 0,width=\widei]{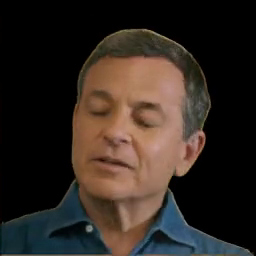} &
    \includegraphics[clip,trim=0 0 0 0,width=\widei]{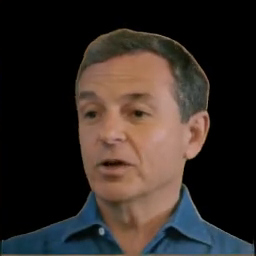} &
    \includegraphics[clip,trim=0 0 0 0,width=\widei]{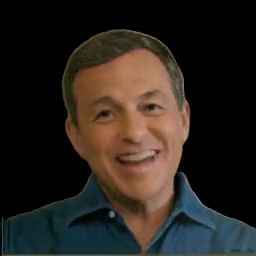} &
    \includegraphics[clip,trim=0 0 0 0,width=\widei]{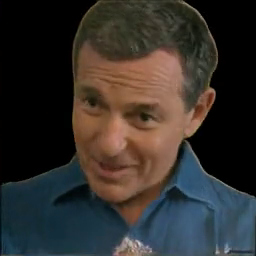} &
    \includegraphics[clip,trim=0 0 0 0,width=\widei]{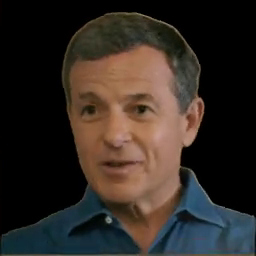} &
    \includegraphics[clip,trim=0 0 0 0,width=\widei]{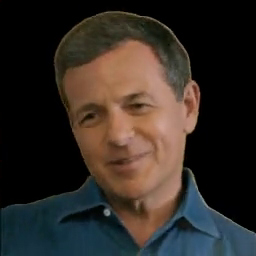} &
    \includegraphics[clip,trim=0 0 0 0,width=\widei]{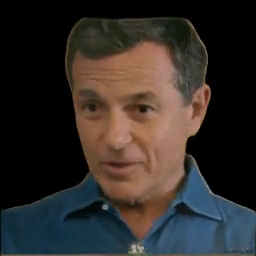} &
    \includegraphics[clip,trim=0 0 0 0,width=\widei]{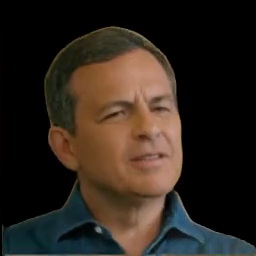} &
    \includegraphics[clip,trim=0 0 0 0,width=\widei]{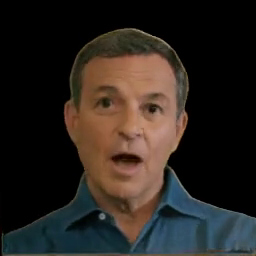} \\
    \end{tabular}
    \caption{Additional comparison of cross-person reenactment for several systems on VoxCeleb2 test set. The layout is the same as in \fig{competitors}.}
    \label{fig:competitors2}
\end{figure*}
\setlength{\tabcolsep}{\oldtabcolsep}

\subsection{Ablation study}
\label{sect:ablation}

In this section, we study the effects of reducing pose vector dimensionality, increasing pose encoder capacity, keeping the background in images, and removing pose augmentation. We retrain our best model with different subsets of these changes, and we also try removing pose augmentations from \textsf{X2Face+} and \textsf{FAb-Net+}. All the resulting models in question are listed in \tab{ablation-models}. We compare them both quantitatively and qualitatively, just as in \sect{reenactment-performance}.

There are four ablation dimensions to explore in our study, and these correspond to the columns in \tab{ablation-models}. We detail and discuss them below using the $\mathcal{I}_T$-vs-$\mathcal{P}_T$ plot in \fig{ablation-quantitative} and the qualitative comparison in \fig{ablation-samples}.

\paragraph{Pose vector dimensionality $d_p$.} First, we try reducing $d_p$ from 256 to 64 simply by changing the number of channels in the last trainable layer of the pose encoder. Our base model (\textsf{Ours}, represented by a red circle) but with pose vectors constrained to 64 dimensions is labeled \textsf{--PoseDim} in \fig{ablation-quantitative}. Intuitively, a tighter bottleneck like this should both limit the ability to represent diverse poses and force the generator to take person-specific information from the richer identity embedding. According to the plot, indeed, the pose reconstruction error increases slightly, while the system stays person-agnostic to a similar degree. Qualitatively, however, the difference in pose is negligible.

\paragraph{Pose encoder capacity.} Second, we attempt replacing the pose encoder with a stronger network, namely ResNeXt-50 (32 $\times$ 4), which makes it even with the identity encoder (that is of the same architecture). We denote our best model with this modification \textsf{+PoseEnc}. As noted in \sect{disentanglement}, in \textsf{Ours} we have intentionally unbalanced the pose and the identity encoder so that the former is weaker, causing the optimization process to favor extracting person-specific information from the identity source frames rather than from the driving frame. Both the metrics and the reenactment samples for \textsf{+PoseEnc} suggest that this idea was not pointless: a more capacious pose encoder starts piping person-specific features from the pose driver. In \fig{ablation-samples}, the \textsf{+PoseEnc}'s result is influenced by clothing from driver \#1, hair from drivers \#6-\#8, facial shape from driver \#9. A huge increase in the identity error in \fig{ablation-quantitative} also confirms this. On the other hand, such system reconstructs the pose with a better accuracy, which may indicate that it might be a better choice for self-reenactment where "identity bleeding" is less of an issue.

\paragraph{Erasing the background.} Third, we modify our system so that it does not predict foreground segmentation, does not use segmentation to compute loss functions, and thus becomes unsupervised. We call \textsf{Ours} plus this change  \textsf{--Segm}. A pose encoder in such system spends its capacity on encoding the driving image's background rather than estimating the subtle details of facial expressions. This happens because the perceptual loss functions are too sensitive to discrepancies between generated and target backgrounds compared to facial expression differences. More importantly, because background often changes within a video, reconstructing the target image's background is too difficult by just looking at the identity source images. Therefore, the optimization algorithm is tempted to offload identity encoder's job to pose encoder. This is evident from the plot and the samples, where introducing backgrounds contributes a lot to the identity gap, even more obviously when combined with a stronger pose encoder (model \textsf{+PoseEnc~--Segm}).

\paragraph{Pose augmentation.} Fourth, we try retraining a model without random pose augmentations, i.e. $A$ is set to an identity transformation. In this setup, a system is trained to \textit{exactly} reconstruct the pose driver image, and is therefore more likely to degrade into an autoencoder (provided the pose encoder is trained along with the whole system). As easy to see from \fig{ablation-quantitative} (\textsf{Ours} $\rightarrow$ \textsf{--Augm},  \textsf{+PoseEnc} $\rightarrow$ \textsf{+PoseEnc~--Augm}), although this further improves the ability to represent poses, it also hurts identity preservation a lot. In fact, a system with a powerful ResNeXt-50 pose encoder trained without pose augmentations (\textsf{+PoseEnc~--Augm}) turned out to be the worst of our models in terms of $\mathcal{P}_T$, but at the same time the best model in terms of pose reconstruction quality. Such a model, again, may be very useful for self-reenactment, but terrible for "puppeteering" (cross-person reenactment). Still, even in self-reenactment, one has to be careful as this model can give undesired effects such as image quality transfer (e.g. from driver \#8 in \fig{ablation-samples}).

This effect is once again confirmed by removing pose augmentations from \textsf{X2Face+} and \textsf{FAb-Net+} (the \textsf{(--Augm)} suffix is added to each model). With random augmentations on, despite the person-specific nature of X2Face and FAb-Net pose descriptors, the generator still develops robustness to person-specific features of pose drivers. Without augmentations, however, the degree of "identity bleeding" becomes fully explained by the identity-specificity of those off-the-shelf descriptors. Also, the pose resonstruction error should decrease given that the generator will not have to be that robust to drivers anymore, and so some of its capacity will free up and may be devoted to render more accurate poses. As expected, \fig{ablation-quantitative} shows severe growth in identity error and a sharp drop in pose error for those two models. This one more time proves that X2Face and FAb-Net descriptors are not person-agnostic. In addition, one can observe the identity gap visually from \fig{ablation-samples-x2face-fabnet}, e.g. how glasses from driver \#7 or facial shape from driver \#1 are transferred to the result. \\

In conclusion, there is a trade-off between identity preservation error $\mathcal{I}_T$ and pose reconstruction error $\mathcal{P}_T$. This trade-off is adjusted by applying the above changes depending on whether the self-reenactment scenario or the cross-person driving scenario is more important. The latter is the case for our best model \textsf{Ours}, while a good candidate for the former setting might be \textsf{+PoseEnc} or \textsf{+PoseEnc~--Segm}.

\setlength{\widei}{0.9\textwidth}
\begin{figure*}
    \centering
    \includegraphics[clip,width=\widei]{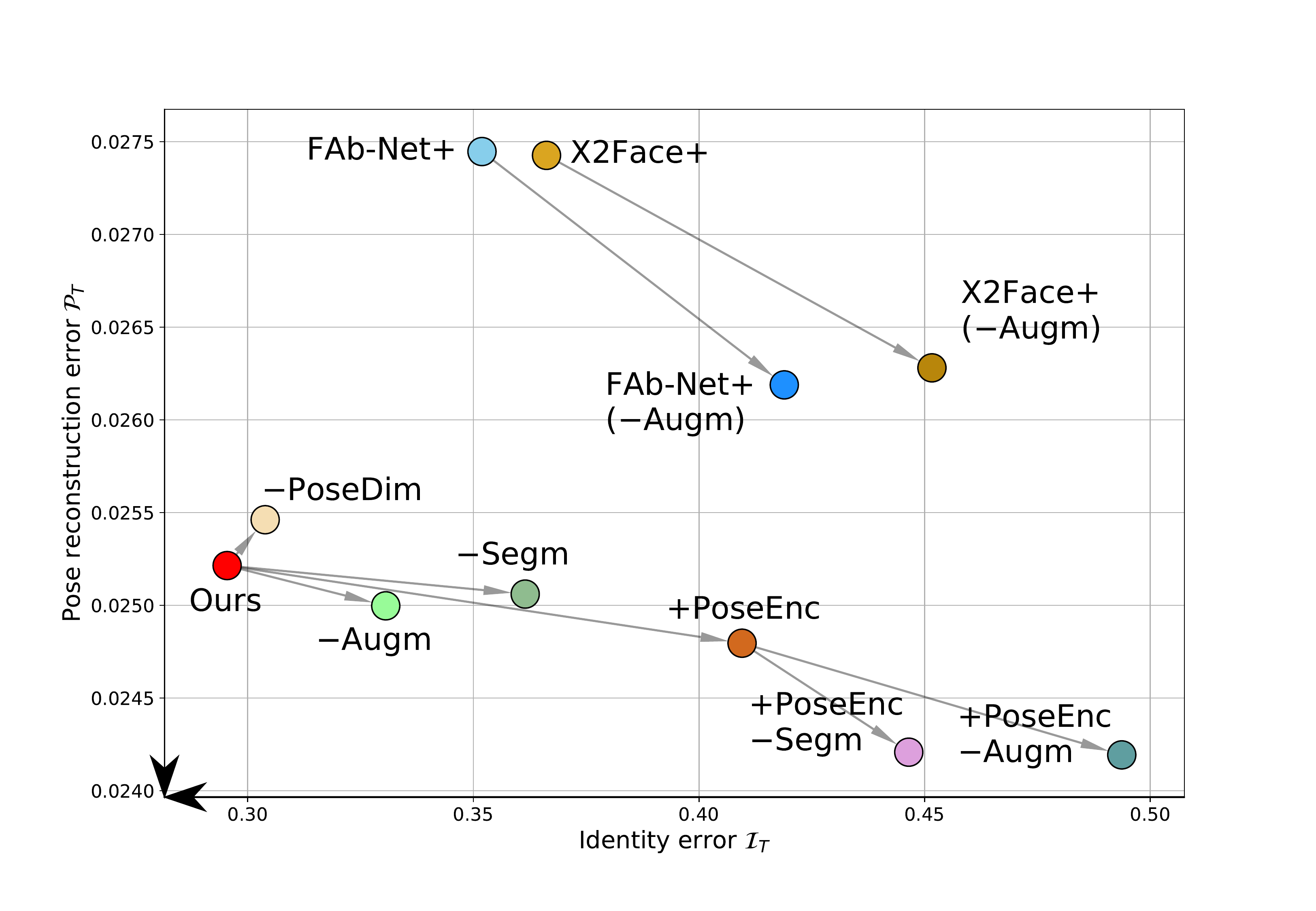}
    \caption{Quantitative evaluation of how ablating several important features of the training setup impacts our system. In addition, the impact of pose augmentation during training is illustrated for \textsf{X2Face+} and \textsf{FAb-Net+}. The legend is given by \tab{ablation-models}. See \sect{ablation} for discussion.}
    \label{fig:ablation-quantitative}
\end{figure*}

\setlength{\oldtabcolsep}{\tabcolsep}
\setlength{\tabcolsep}{5pt}
\begin{table*}
\centering
\begin{tabular}{l|c|c|c|c}
\makecell{\textbf{Model name}} & \makecell{\textbf{Pose vector} \\ \textbf{dim. $d_p$}} & \makecell{\textbf{Pose} \\ \textbf{encoder}} & \makecell{\textbf{Erased} \\ \textbf{background}} & \makecell{\textbf{Pose} \\ \textbf{augmentation}} \\ \toprule[1.5pt]
\textsf{Ours} & 256 & \multirow{4}{*}{\makecell[c]{MobileNetV2}} & + & + \\ \cline{1-2}\cline{4-5}
\textsf{--PoseDim} & 64 &  & + & + \\ \cline{1-2}\cline{4-5}
\textsf{--Augm} & 256 &  & + & \\ \cline{1-2}\cline{4-5}
\textsf{--Segm} & 256 &  &  & + \\ \hline
\textsf{+PoseEnc} & 256 & \multirow{3}{*}{\makecell{ResNeXt-50 \\ (32 $\times$ 4d)}} & + & + \\ \cline{1-2}\cline{4-5}
\textsf{+PoseEnc~--Augm} & 256 &  & + & \\ \cline{1-2}\cline{4-5}
\textsf{+PoseEnc~--Segm} & 256 &  &  & + \\ \toprule[1.1pt]
\textsf{X2Face+} & 128 & \multirow{2}{*}{\makecell{X2Face \\ (pretrained)}} & + & + \\ \cline{1-2}\cline{4-5}
\textsf{X2Face+ (--Augm)} & 128 &    & + & \\ \hline
\textsf{FAb-Net+} & 256 & \multirow{2}{*}{\makecell{FAb-Net \\ (pretrained)}} & + & + \\ \cline{1-2}\cline{4-5}
\textsf{FAb-Net+ (--Augm)} & 256 &  & + & 
\end{tabular}
\caption{A summary of systems compared in the ablation study.}
\label{tab:ablation-models}
\end{table*}
\setlength{\tabcolsep}{\oldtabcolsep}

\setlength{\oldtabcolsep}{\tabcolsep}
\setlength{\tabcolsep}{1pt}
\setlength{\widei}{0.0984\textwidth}
\begin{figure*}
    \centering
    \begin{tabular}{M{\widei}M{\widei}M{\widei}M{\widei}M{\widei}M{\widei}M{\widei}M{\widei}M{\widei}M{\widei}}
    \includegraphics[clip,width=\widei]{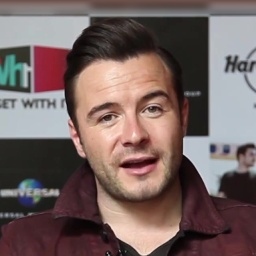} &
    \includegraphics[clip,trim=0 0 0 0,width=\widei]{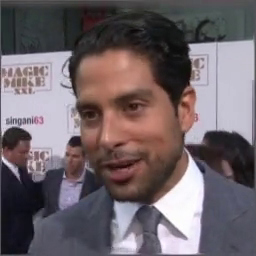} &
    \includegraphics[clip,trim=0 0 0 0,width=\widei]{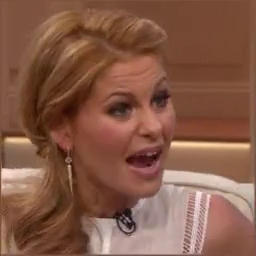} &
    \includegraphics[clip,trim=0 0 0 0,width=\widei]{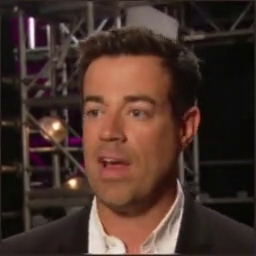} &
    \includegraphics[clip,trim=0 0 0 0,width=\widei]{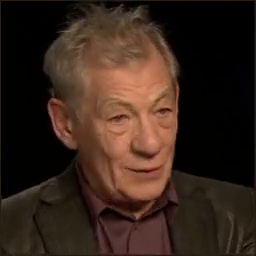} &
    \includegraphics[clip,trim=0 0 0 0,width=\widei]{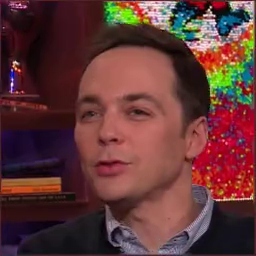} &
    \includegraphics[clip,trim=0 0 0 0,width=\widei]{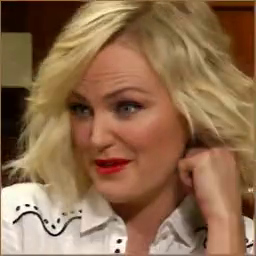} &
    \includegraphics[clip,trim=0 0 0 0,width=\widei]{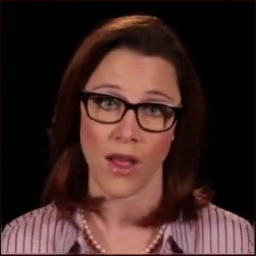} &
    \includegraphics[clip,trim=0 0 0 0,width=\widei]{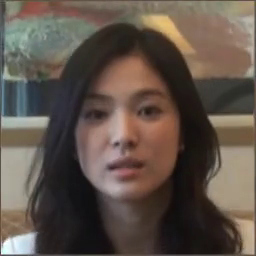} &
    \includegraphics[clip,trim=0 0 0 0,width=\widei]{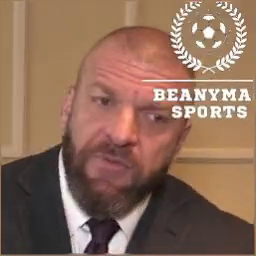} \\
    \textbf{\textsf{Ours}} &
    \includegraphics[clip,trim=0 0 0 0,width=\widei]{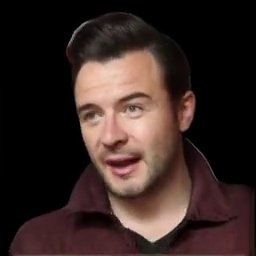} &
    \includegraphics[clip,trim=0 0 0 0,width=\widei]{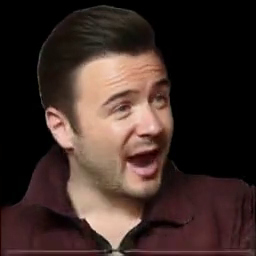} &
    \includegraphics[clip,trim=0 0 0 0,width=\widei]{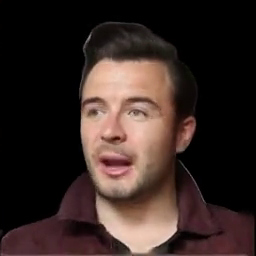} &
    \includegraphics[clip,trim=0 0 0 0,width=\widei]{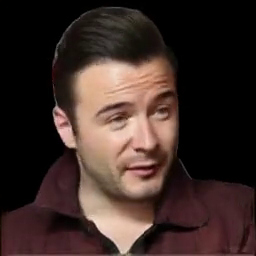} &
    \includegraphics[clip,trim=0 0 0 0,width=\widei]{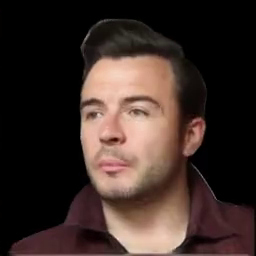} &
    \includegraphics[clip,trim=0 0 0 0,width=\widei]{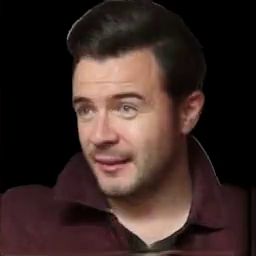} &
    \includegraphics[clip,trim=0 0 0 0,width=\widei]{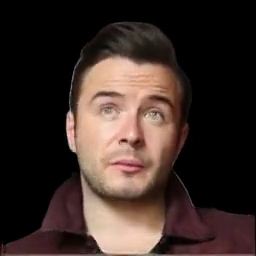} &
    \includegraphics[clip,trim=0 0 0 0,width=\widei]{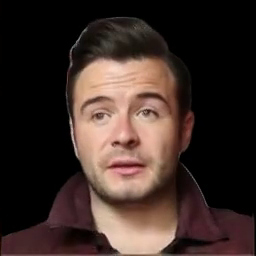} &
    \includegraphics[clip,trim=0 0 0 0,width=\widei]{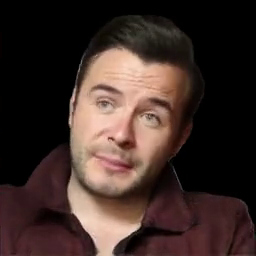} \\
    \textbf{\textsf{--PoseDim}} &
    \includegraphics[clip,trim=0 0 0 0,width=\widei]{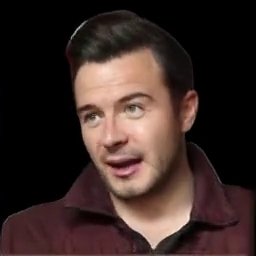} &
    \includegraphics[clip,trim=0 0 0 0,width=\widei]{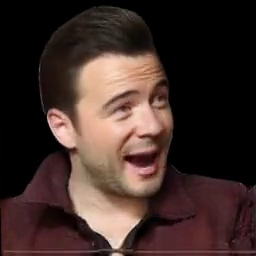} &
    \includegraphics[clip,trim=0 0 0 0,width=\widei]{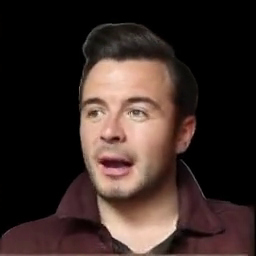} &
    \includegraphics[clip,trim=0 0 0 0,width=\widei]{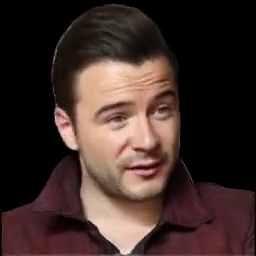} &
    \includegraphics[clip,trim=0 0 0 0,width=\widei]{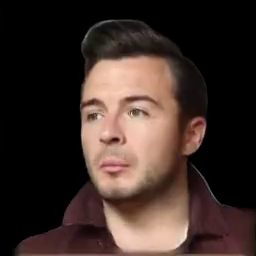} &
    \includegraphics[clip,trim=0 0 0 0,width=\widei]{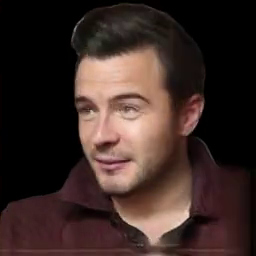} &
    \includegraphics[clip,trim=0 0 0 0,width=\widei]{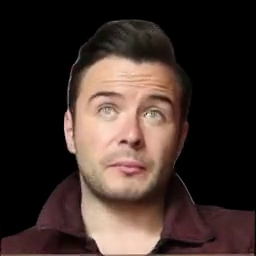} &
    \includegraphics[clip,trim=0 0 0 0,width=\widei]{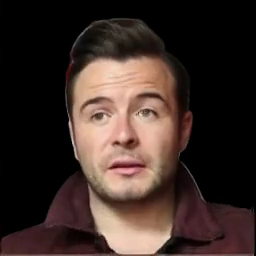} &
    \includegraphics[clip,trim=0 0 0 0,width=\widei]{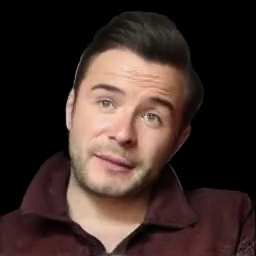} \\
    \textbf{\textsf{--Augm}} &
    \includegraphics[clip,trim=0 0 0 0,width=\widei]{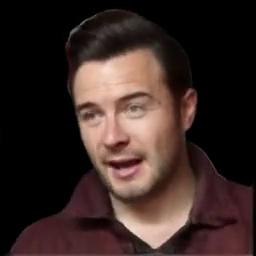} &
    \includegraphics[clip,trim=0 0 0 0,width=\widei]{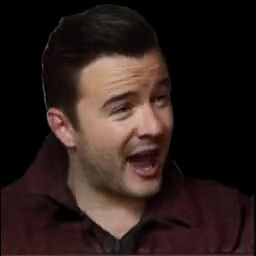} &
    \includegraphics[clip,trim=0 0 0 0,width=\widei]{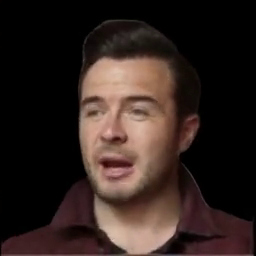} &
    \includegraphics[clip,trim=0 0 0 0,width=\widei]{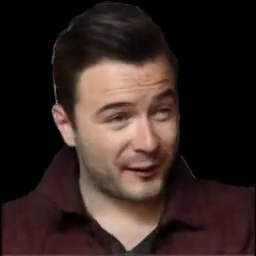} &
    \includegraphics[clip,trim=0 0 0 0,width=\widei]{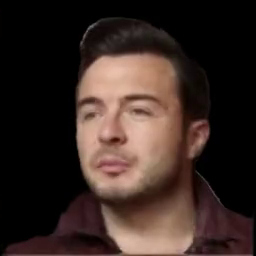} &
    \includegraphics[clip,trim=0 0 0 0,width=\widei]{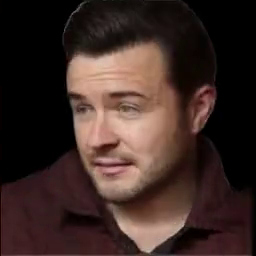} &
    \includegraphics[clip,trim=0 0 0 0,width=\widei]{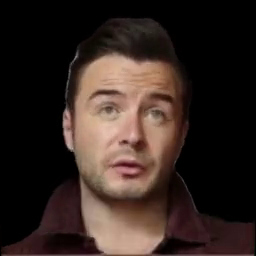} &
    \includegraphics[clip,trim=0 0 0 0,width=\widei]{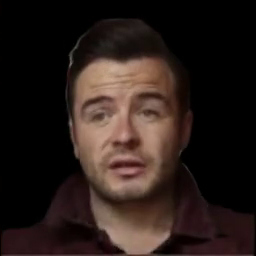} &
    \includegraphics[clip,trim=0 0 0 0,width=\widei]{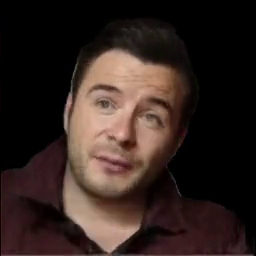} \\
    \textbf{\textsf{--Segm}} &
    \includegraphics[clip,trim=0 0 0 0,width=\widei]{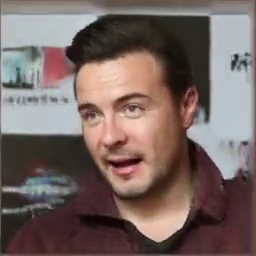} &
    \includegraphics[clip,trim=0 0 0 0,width=\widei]{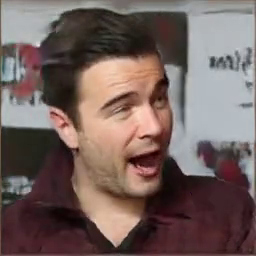} &
    \includegraphics[clip,trim=0 0 0 0,width=\widei]{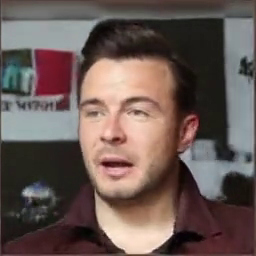} &
    \includegraphics[clip,trim=0 0 0 0,width=\widei]{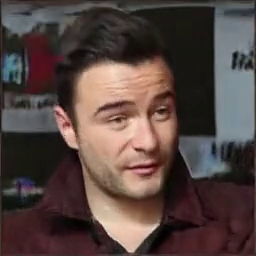} &
    \includegraphics[clip,trim=0 0 0 0,width=\widei]{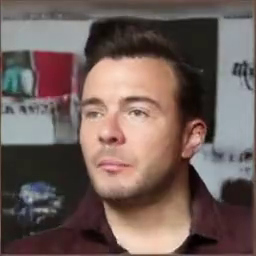} &
    \includegraphics[clip,trim=0 0 0 0,width=\widei]{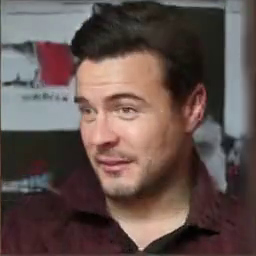} &
    \includegraphics[clip,trim=0 0 0 0,width=\widei]{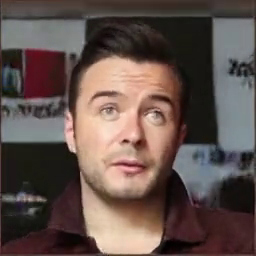} &
    \includegraphics[clip,trim=0 0 0 0,width=\widei]{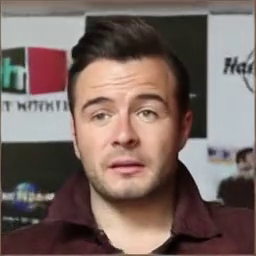} &
    \includegraphics[clip,trim=0 0 0 0,width=\widei]{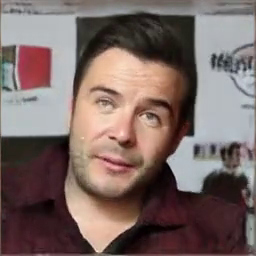} \\
    \textbf{\textsf{+PoseEnc}} &
    \includegraphics[clip,trim=0 0 0 0,width=\widei]{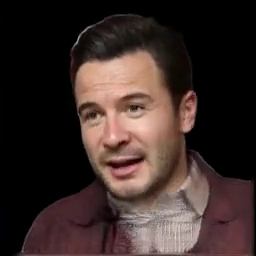} &
    \includegraphics[clip,trim=0 0 0 0,width=\widei]{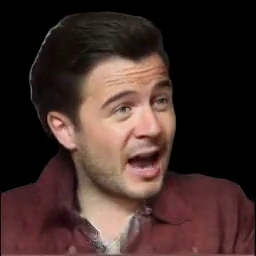} &
    \includegraphics[clip,trim=0 0 0 0,width=\widei]{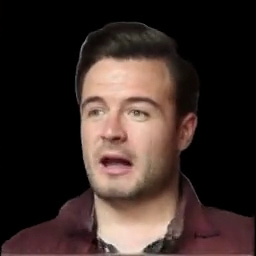} &
    \includegraphics[clip,trim=0 0 0 0,width=\widei]{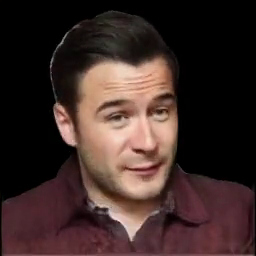} &
    \includegraphics[clip,trim=0 0 0 0,width=\widei]{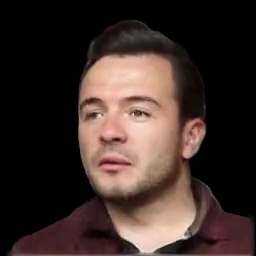} &
    \includegraphics[clip,trim=0 0 0 0,width=\widei]{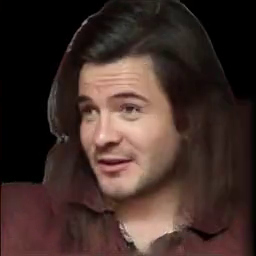} &
    \includegraphics[clip,trim=0 0 0 0,width=\widei]{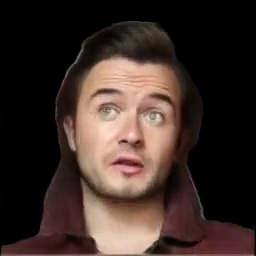} &
    \includegraphics[clip,trim=0 0 0 0,width=\widei]{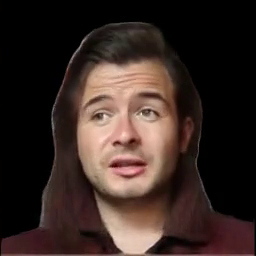} &
    \includegraphics[clip,trim=0 0 0 0,width=\widei]{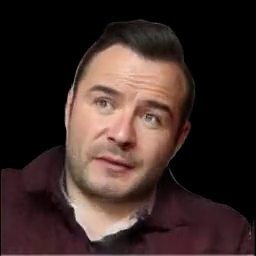} \\
    \makecell{\textbf{\textsf{+PoseEnc}} \\ \textbf{\textsf{--Augm}}} &
    \includegraphics[clip,trim=0 0 0 0,width=\widei]{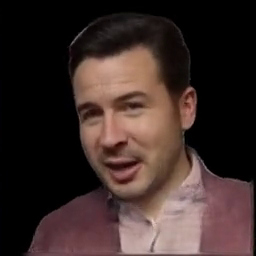} &
    \includegraphics[clip,trim=0 0 0 0,width=\widei]{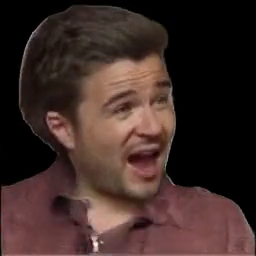} &
    \includegraphics[clip,trim=0 0 0 0,width=\widei]{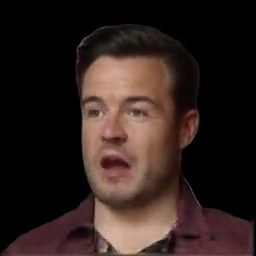} &
    \includegraphics[clip,trim=0 0 0 0,width=\widei]{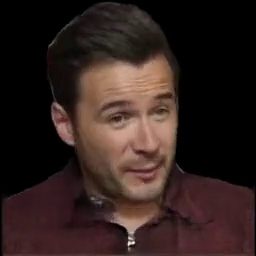} &
    \includegraphics[clip,trim=0 0 0 0,width=\widei]{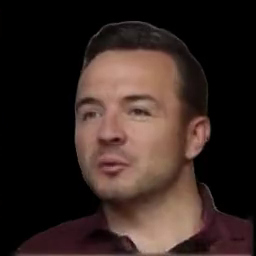} &
    \includegraphics[clip,trim=0 0 0 0,width=\widei]{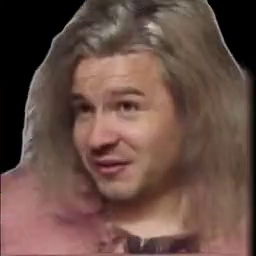} &
    \includegraphics[clip,trim=0 0 0 0,width=\widei]{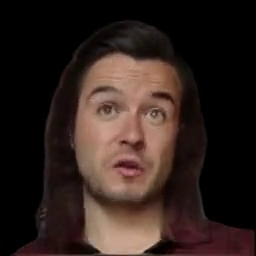} &
    \includegraphics[clip,trim=0 0 0 0,width=\widei]{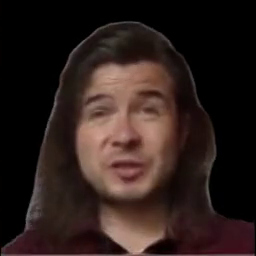} &
    \includegraphics[clip,trim=0 0 0 0,width=\widei]{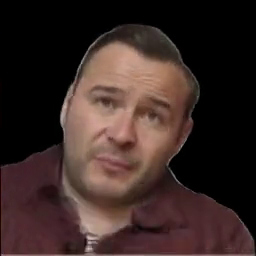} \\
    \makecell{\textbf{\textsf{+PoseEnc}} \\ \textbf{\textsf{--Segm}}} &
    \includegraphics[clip,trim=0 0 0 0,width=\widei]{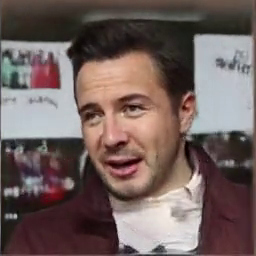} &
    \includegraphics[clip,trim=0 0 0 0,width=\widei]{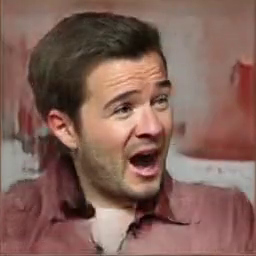} &
    \includegraphics[clip,trim=0 0 0 0,width=\widei]{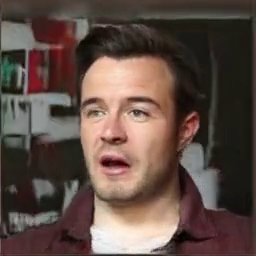} &
    \includegraphics[clip,trim=0 0 0 0,width=\widei]{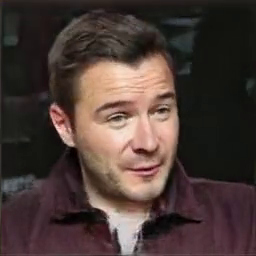} &
    \includegraphics[clip,trim=0 0 0 0,width=\widei]{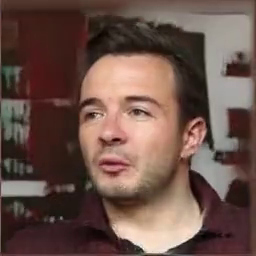} &
    \includegraphics[clip,trim=0 0 0 0,width=\widei]{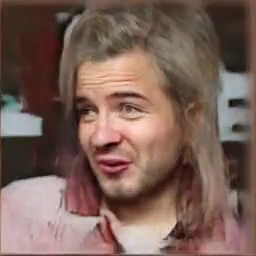} &
    \includegraphics[clip,trim=0 0 0 0,width=\widei]{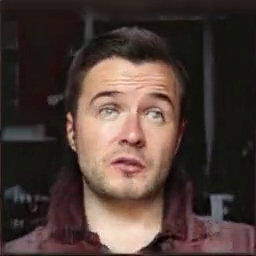} &
    \includegraphics[clip,trim=0 0 0 0,width=\widei]{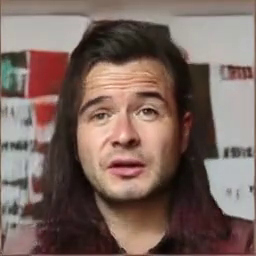} &
    \includegraphics[clip,trim=0 0 0 0,width=\widei]{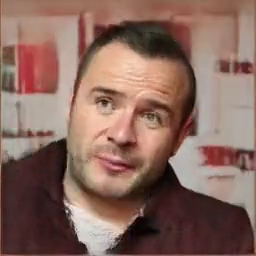} \\
    \end{tabular}
    \caption{Comparison of cross-person reenactment for our best model and its ablated versions. See \sect{ablation} for discussion.}
    \label{fig:ablation-samples}
\end{figure*}
\setlength{\tabcolsep}{\oldtabcolsep}

\setlength{\oldtabcolsep}{\tabcolsep}
\setlength{\tabcolsep}{1pt}
\setlength{\widei}{0.0984\textwidth}
\begin{figure*}
    \centering
    \begin{tabular}{M{\widei}M{\widei}M{\widei}M{\widei}M{\widei}M{\widei}M{\widei}M{\widei}M{\widei}M{\widei}}
\includegraphics[clip,width=\widei]{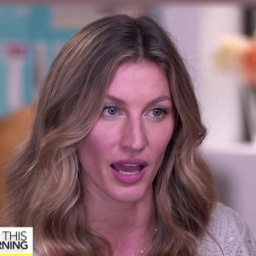} &
    \includegraphics[clip,trim=0 0 0 0,width=\widei]{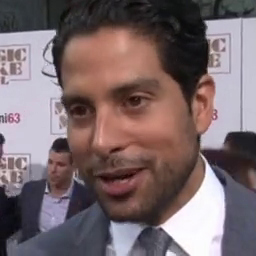} &
    \includegraphics[clip,trim=0 0 0 0,width=\widei]{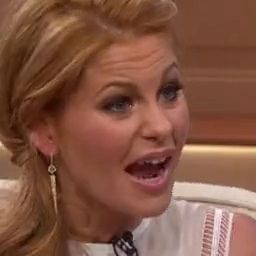} &
    \includegraphics[clip,trim=0 0 0 0,width=\widei]{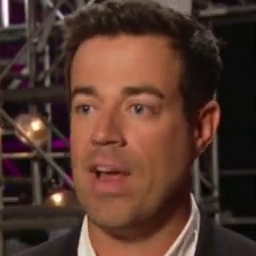} &
    \includegraphics[clip,trim=0 0 0 0,width=\widei]{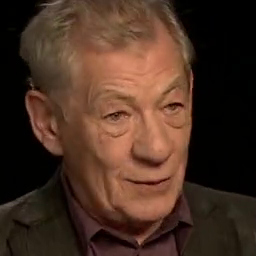} &
    \includegraphics[clip,trim=0 0 0 0,width=\widei]{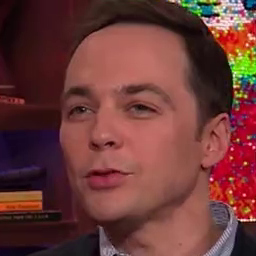} &
    \includegraphics[clip,trim=0 0 0 0,width=\widei]{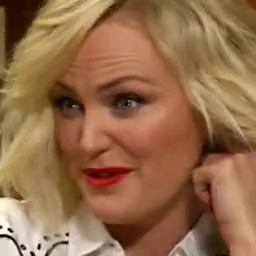} &
    \includegraphics[clip,trim=0 0 0 0,width=\widei]{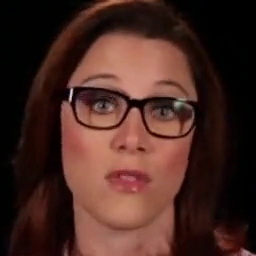} &
    \includegraphics[clip,trim=0 0 0 0,width=\widei]{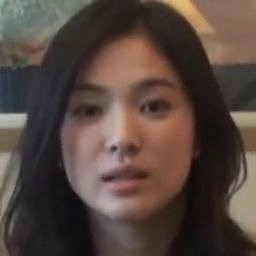} &
    \includegraphics[clip,trim=0 0 0 0,width=\widei]{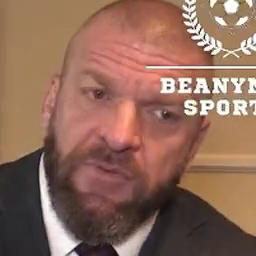} \\
    \makecell{\textbf{\textsf{X2Face+}}} &
    \includegraphics[clip,trim=0 0 0 0,width=\widei]{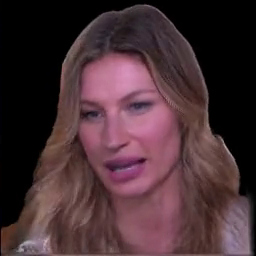} &
    \includegraphics[clip,trim=0 0 0 0,width=\widei]{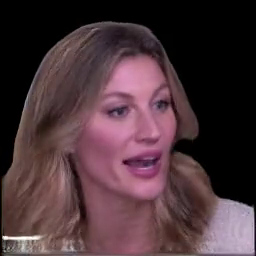} &
    \includegraphics[clip,trim=0 0 0 0,width=\widei]{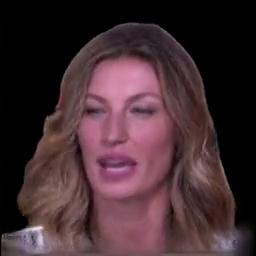} &
    \includegraphics[clip,trim=0 0 0 0,width=\widei]{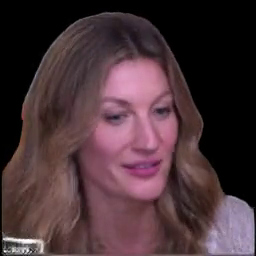} &
    \includegraphics[clip,trim=0 0 0 0,width=\widei]{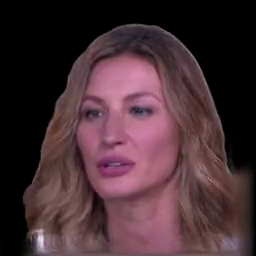} &
    \includegraphics[clip,trim=0 0 0 0,width=\widei]{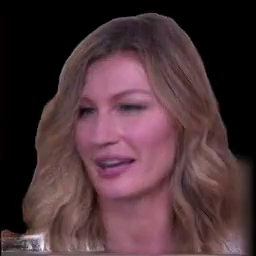} &
    \includegraphics[clip,trim=0 0 0 0,width=\widei]{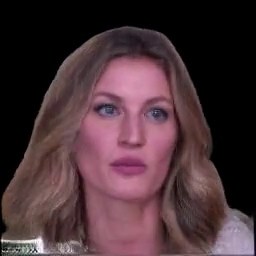} &
    \includegraphics[clip,trim=0 0 0 0,width=\widei]{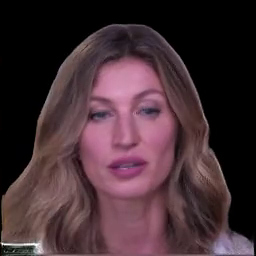} &
    \includegraphics[clip,trim=0 0 0 0,width=\widei]{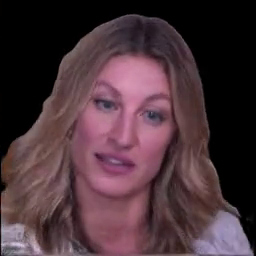} \\
    \makecell{\textbf{\textsf{X2Face+}} \\ \textbf{\textsf{(--Augm)}}} &
    \includegraphics[clip,trim=0 0 0 0,width=\widei]{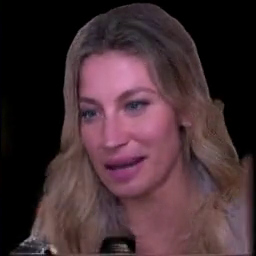} &
    \includegraphics[clip,trim=0 0 0 0,width=\widei]{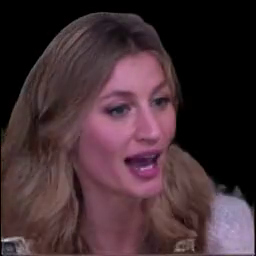} &
    \includegraphics[clip,trim=0 0 0 0,width=\widei]{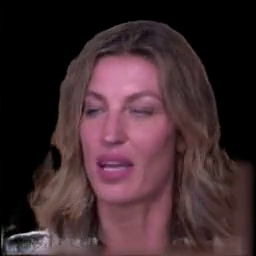} &
    \includegraphics[clip,trim=0 0 0 0,width=\widei]{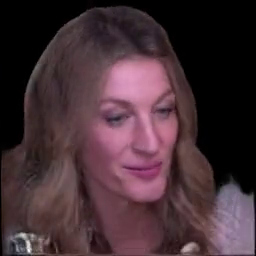} &
    \includegraphics[clip,trim=0 0 0 0,width=\widei]{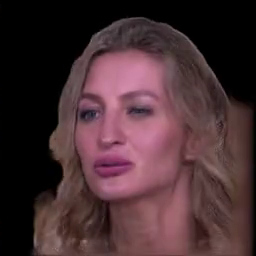} &
    \includegraphics[clip,trim=0 0 0 0,width=\widei]{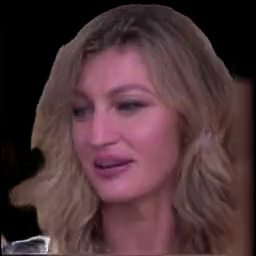} &
    \includegraphics[clip,trim=0 0 0 0,width=\widei]{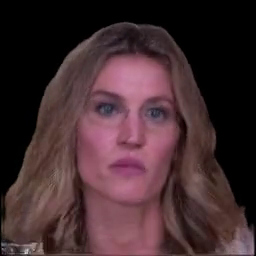} &
    \includegraphics[clip,trim=0 0 0 0,width=\widei]{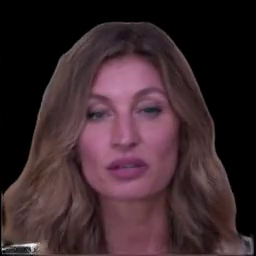} &
    \includegraphics[clip,trim=0 0 0 0,width=\widei]{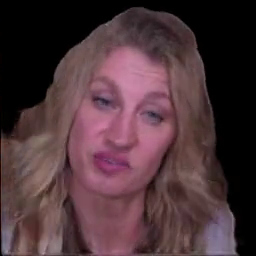} \\
    \makecell{\textbf{\textsf{FAb-Net+}}} &
    \includegraphics[clip,trim=0 0 0 0,width=\widei]{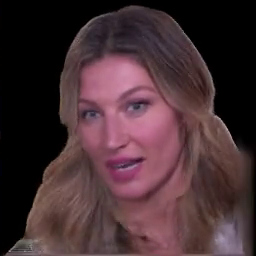} &
    \includegraphics[clip,trim=0 0 0 0,width=\widei]{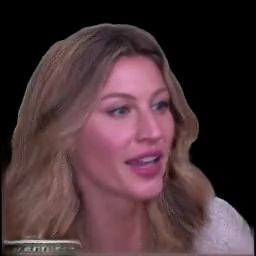} &
    \includegraphics[clip,trim=0 0 0 0,width=\widei]{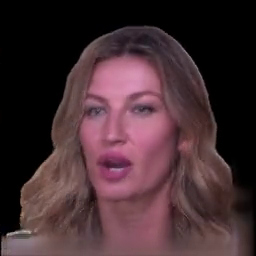} &
    \includegraphics[clip,trim=0 0 0 0,width=\widei]{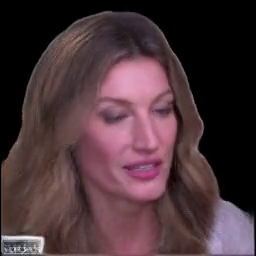} &
    \includegraphics[clip,trim=0 0 0 0,width=\widei]{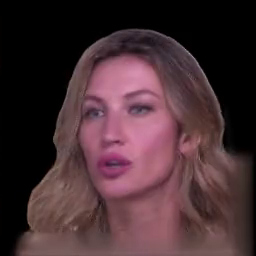} &
    \includegraphics[clip,trim=0 0 0 0,width=\widei]{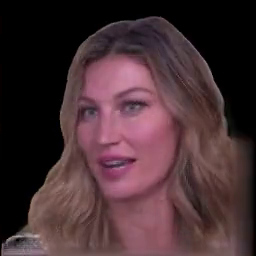} &
    \includegraphics[clip,trim=0 0 0 0,width=\widei]{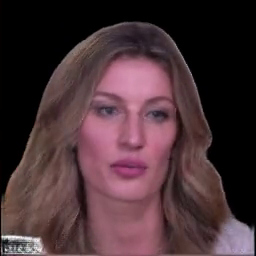} &
    \includegraphics[clip,trim=0 0 0 0,width=\widei]{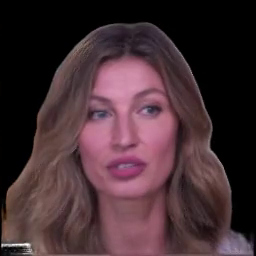} &
    \includegraphics[clip,trim=0 0 0 0,width=\widei]{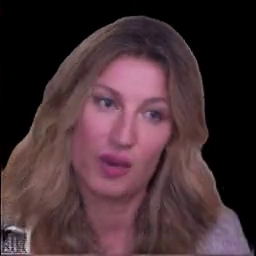} \\
    \makecell{\textbf{\textsf{FAb-Net+}} \\ \textbf{\textsf{(--Augm)}}} &
    \includegraphics[clip,trim=0 0 0 0,width=\widei]{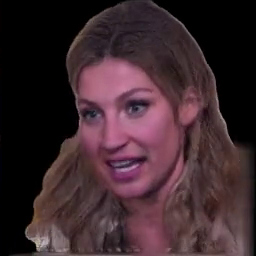} &
    \includegraphics[clip,trim=0 0 0 0,width=\widei]{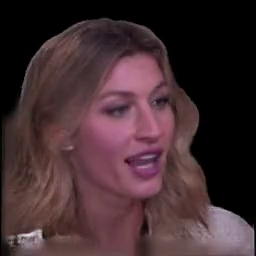} &
    \includegraphics[clip,trim=0 0 0 0,width=\widei]{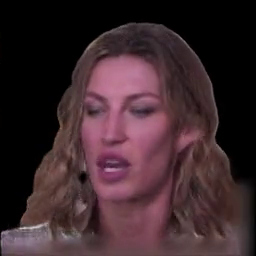} &
    \includegraphics[clip,trim=0 0 0 0,width=\widei]{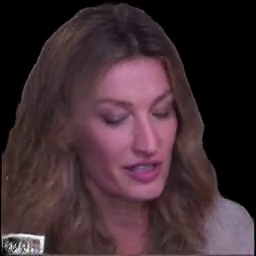} &
    \includegraphics[clip,trim=0 0 0 0,width=\widei]{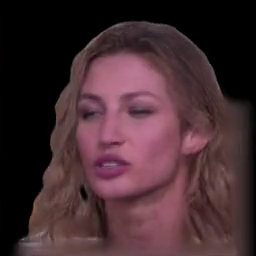} &
    \includegraphics[clip,trim=0 0 0 0,width=\widei]{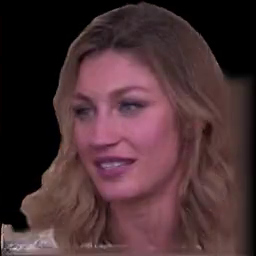} &
    \includegraphics[clip,trim=0 0 0 0,width=\widei]{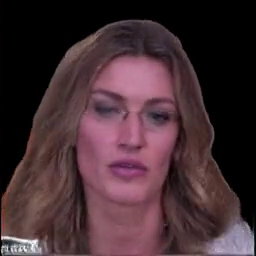} &
    \includegraphics[clip,trim=0 0 0 0,width=\widei]{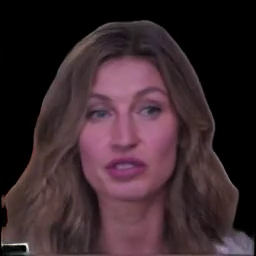} &
    \includegraphics[clip,trim=0 0 0 0,width=\widei]{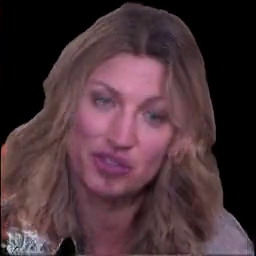}
    \end{tabular}
    \caption{The effect of pose augmentations on \textsf{X2Face+} and \textsf{FAb-Net+} models. Without augmentations, the identity gap becomes conspicuous.}
    \label{fig:ablation-samples-x2face-fabnet}
\end{figure*}
\setlength{\tabcolsep}{\oldtabcolsep}

\end{document}